%% file: access.tex
\documentclass{ieeeaccess}
\usepackage{cite}
\usepackage{amsmath,amssymb,amsfonts}
\usepackage{algorithmic}
\usepackage{graphicx}
\usepackage{textcomp}
\usepackage{glossaries}
\usepackage{bm}
\usepackage{enumitem}
\usepackage{subcaption}
\usepackage{multirow}
\usepackage{float}
\makeatletter
\AtBeginDocument{\DeclareMathVersion{bold}
\SetSymbolFont{operators}{bold}{T1}{times}{b}{n}
\SetSymbolFont{NewLetters}{bold}{T1}{times}{b}{it}
\SetMathAlphabet{\mathrm}{bold}{T1}{times}{b}{n}
\SetMathAlphabet{\mathit}{bold}{T1}{times}{b}{it}
\SetMathAlphabet{\mathbf}{bold}{T1}{times}{b}{n}
\SetMathAlphabet{\mathtt}{bold}{OT1}{pcr}{b}{n}
\SetSymbolFont{symbols}{bold}{OMS}{cmsy}{b}{n}
\renewcommand\boldmath{\@nomath\boldmath\mathversion{bold}}}
\makeatother

\def\BibTeX{{\rm B\kern-.05em{\sc i\kern-.025em b}\kern-.08em
    T\kern-.1667em\lower.7ex\hbox{E}\kern-.125emX}}

\newcommand{\x}{\ifmmode\bm{x}\else\textbf{x}\fi}
\newcommand{\y}{\ifmmode\bm{y}\else\textbf{y}\fi}
\newcommand{\z}{\ifmmode\bm{z}\else\textbf{z}\fi}
\newcommand{\C}{\ifmmode\bm{c}\else\textbf{c}\fi}
\newcommand{\K}{\ifmmode\bm{K}\else\textbf{K}\fi}

\begin{document}
\input{glossary}

\history{Date of publication xxxx 00, 0000, date of current version xxxx 00, 0000.}
\doi{10.1109/ACCESS.2024.0429000}

\title{Point Cloud Geometry Scalable Coding Using a Resolution and Quality-conditioned Latents Probability Estimator}
\author{\uppercase{Daniele Mari}\authorrefmark{1},
\uppercase{André F. R. Guarda}\authorrefmark{2}, \IEEEmembership{Member, IEEE}, \uppercase{Nuno M. M. Rodrigues}\authorrefmark{2}\authorrefmark{4}, \IEEEmembership{Senior Member, IEEE}, \uppercase{Simone Milani}\authorrefmark{1}, \IEEEmembership{Member, IEEE}, and \uppercase{Fernando Pereira}\authorrefmark{2}\authorrefmark{3}, \IEEEmembership{Fellow, IEEE}}

\address[1]{University of Padova, Department of Information Engineering, Padova, 35131 Italy (e-mail: name.surname@dei.unipd.it)}
\address[2]{Instituto de Telecomunicações, 1049-001 Lisbon, Portugal}
\address[3]{Instituto Superior Técnico, Universidade de Lisboa, 1049-001 Lisbon, Portugal)}
\address[4]{ESTG, Politécnico de Leiria, 2411-901 Leiria, Portugal}
\tfootnote{This work was partially supported by the European Union under the Italian National Recovery and Resilience Plan (NRRP) of NextGenerationEU, with a partnership on “Telecommunications of the Future” (PE00000001 - program “RESTART”). This study was also carried out within the Future Artificial Intelligence Research (FAIR) and received funding from the European Union Next-GenerationEU (PNRR - Piano Nazionale di Ripresa e Resilienza - Missione 4, Componente 2, Investimento 1.3 - D.D. 1555 11/10/2022, PE00000013). This work was also funded by the Fundação para a Ciência e a Tecnologia (FCT, Portugal) through the research project PTDC/EEI-COM/1125/2021, entitled “Deep Learning-based Point Cloud Representation”, and by FCT/MECI through national funds and when applicable co-funded EU funds under UID/50008: Instituto de Telecomunicações. This manuscript reflects only the authors’ views and opinions, neither the EU nor the European Commission can be considered responsible for them. Daniele Mari's activities were supported by Fondazione CaRiPaRo under the grants “Dottorati di Ricerca” 2021/2022.}

\markboth
{D. Mari \headeretal: Preparation of Papers for IEEE TRANSACTIONS and JOURNALS}
{D. Mari \headeretal: Preparation of Papers for IEEE TRANSACTIONS and JOURNALS}

\corresp{Corresponding author: Daniele Mari (e-mail: daniele.mari@dei.unipd.it).}

\begin{abstract}
In the current age, users consume multimedia content in very heterogeneous scenarios in terms of network, hardware, and display capabilities. A naive solution to this problem is to encode multiple independent streams, each covering a different possible requirement for the clients, with an obvious negative impact in both storage and computational requirements. These drawbacks can be avoided by using codecs that enable scalability, i.e., the ability to generate a progressive bitstream, containing a base layer followed by multiple enhancement layers, that allow decoding the same bitstream serving multiple reconstructions and visualization specifications.
While scalable coding is a well-known and addressed feature in conventional image and video codecs, this paper focuses on a new and very different problem, notably the development of scalable coding solutions for deep learning-based \gls{pc} coding. The peculiarities of this 3D representation make it hard to implement flexible solutions that do not compromise the other functionalities of the codec.
This paper proposes a joint quality and resolution scalability scheme, named \gls{esqh}, that, contrary to previous solutions, can model the relationship between latents obtained with models trained for different RD tradeoffs and/or at different resolutions.
Experimental results obtained by integrating \gls{esqh} in the emerging JPEG Pleno learning-based PC coding standard show that \gls{esqh}  allows decoding the PC at different qualities and resolutions with a single bitstream while incurring only in a limited RD penalty and increment in complexity w.r.t. non-scalable JPEG PCC that would require one bitstream per coding configuration.
\end{abstract}


\titlepgskip=-21pt

\maketitle

\input{01_intro}
\input{02_previous_works}
\input{03_vm}
\input{04_method}

\input{05_results}
\input{06_conclusions}
\section*{Acknowledgment}
After the first draft, the text in the various sections of the manuscript (except the Abstract) was improved using  Claude 3.5 Sonnet through the Perplexity AI web app. After this process, the paper underwent multiple sets of reviews from the authors and was thus further improved and modified.
\bibliographystyle{IEEEtran}
\bibliography{biblio}

\begin{IEEEbiography}[{\includegraphics[width=1in,height=1.25in,clip,keepaspectratio]{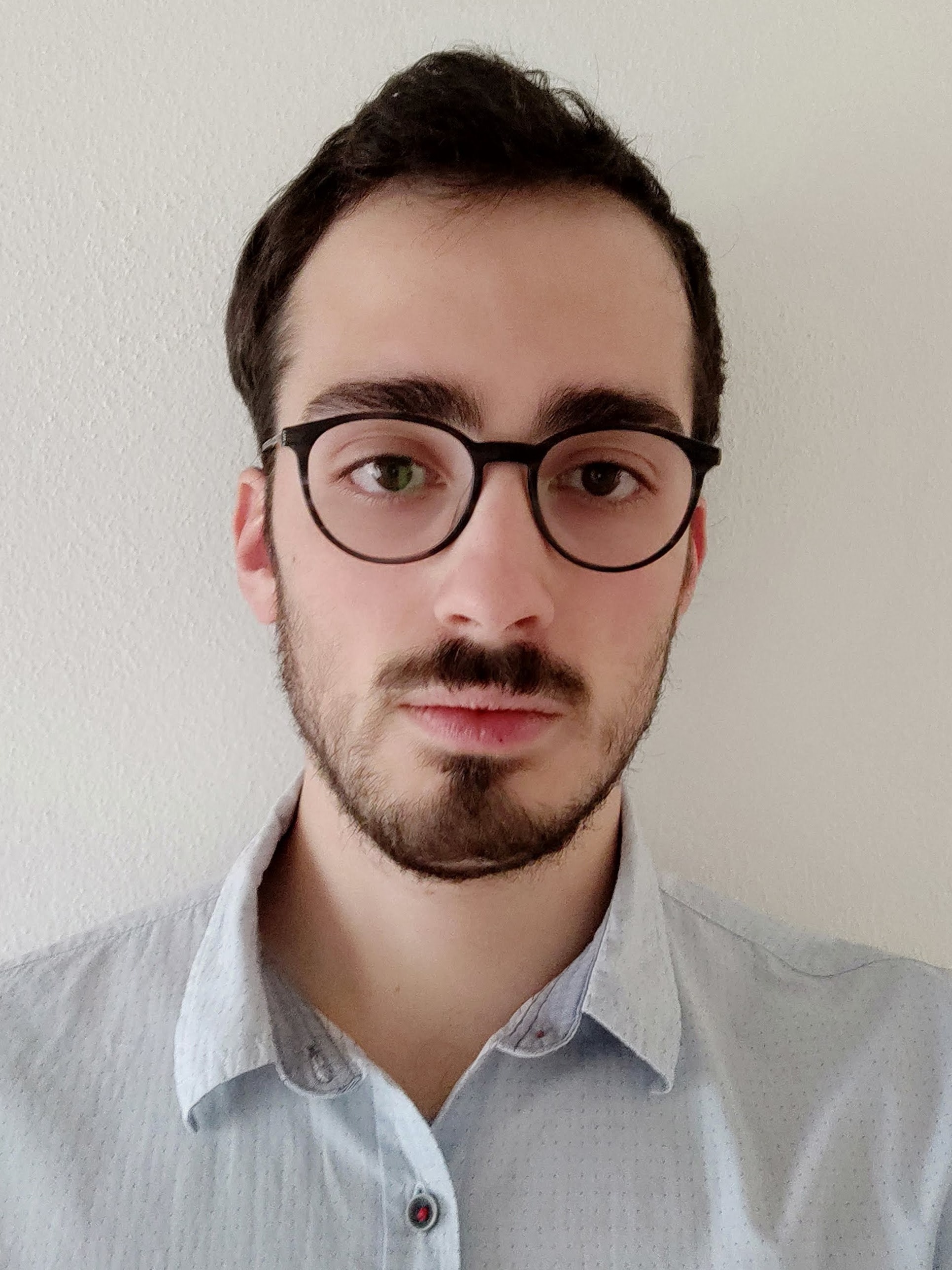}}]{Daniele Mari} received the B.S. degree in information engineering and M.S. degree in ICT for Internet and Multimedia from the University of Padova, Italy, in 2019 and 2021 respectively. He is currently pursuing his Ph.D. in information engineering  and is currently a Ph.D. student in Padova since 2021. Additionally, he has spent 6 months as a visiting Ph.D. student in Instituto Superior Técnico, Universidade de Lisboa, Portugal, in 2023. His main research interests are learned point cloud and image coding. He has authored several publications in top conferences and journals in this field.
\end{IEEEbiography}

\begin{IEEEbiography}[{\includegraphics[width=1in,height=1.25in,clip,keepaspectratio]{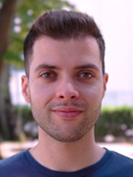}}]{ANDRÉ F. R. GUARDA} (Member, IEEE) received his B.Sc. and M.Sc. degrees in electrotechnical engineering from Instituto Politécnico de Leiria, Portugal, in 2013 and 2016, respectively, and the Ph.D. degree in electrical and computer engineering from Instituto Superior Técnico, Universidade de Lisboa, Portugal, in 2021. He has been a researcher at Instituto de Telecomunicações since 2011, where he currently holds a Post-Doctoral position. His main research interests include multimedia signal processing and coding, with particular focus on point cloud coding with deep learning. He has authored several publications in top conferences and journals in this field and is actively contributing to the standardization efforts of JPEG and MPEG on learning-based point cloud coding.
\end{IEEEbiography}

\newpage


\begin{IEEEbiography}[{\includegraphics[width=1in,height=1.25in,clip,keepaspectratio]{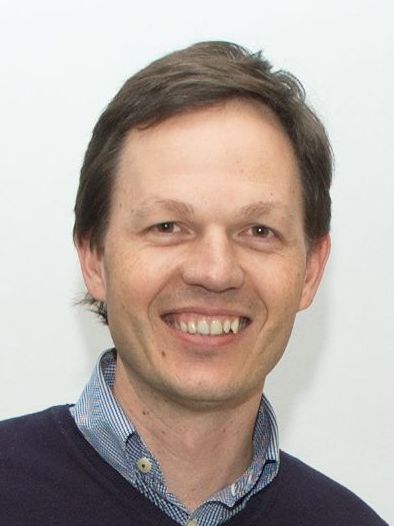}}]{NUNO M. M. RODRIGUES} 
 (Senior Member,
IEEE) graduated in electrical engineering in 1997, received the M.Sc. degree from the Universidade de Coimbra, Portugal, in 2000, and the Ph.D. degree from the Universidade de Coimbra, Portugal, in 2009, in collaboration with the Universidade Federal do Rio de Janeiro, Brazil. He is a Professor in the Department of Electrical Engineering, in the School of Technology and Management of the Polytechnic University of Leiria, Portugal and a Senior Researcher in Instituto de Telecomunicações, Portugal. He has coordinated and participated as a researcher in various national and international funded projects. He has supervised three concluded PhD theses and several MSc theses. He is co-author of a book and more than 100 publications, including book chapters and papers in national and international journals and conferences. His research interests include several topics related with image and video coding and processing, for different signal modalities and applications. His current research is focused on deep learning-based techniques for point cloud coding and processing.
\end{IEEEbiography}

\begin{IEEEbiography}[{\includegraphics[width=1in,height=1.25in,clip,keepaspectratio]{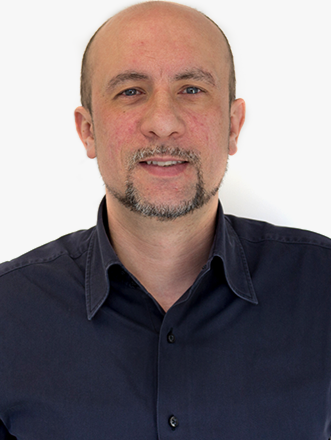}}]{Simone Milani} 
(Member, IEEE) received the
Laurea degree in telecommunication engineering
and the Ph.D. degree in electronics and telecommunication engineering from the University of
Padova, Padova, Italy, in 2002 and 2007, respectively. He was a Visiting Ph.D. Student at the University of California at Berkeley, Berkeley, CA,
USA, in 2006. He was a Consultant at STMicroelectronics, Agrate, Italy. He was a Postdoctoral Researcher at the University of Udine, Udine,
Italy, the University of Padova, and the Politecnico di Milano, Milan, Italy,
from 2007 to 2013. From 2013 to 2020, he was an Assistant Professor with
the Department of Information Engineering, University of Padova, where
he is an Associate Professor. His research interests include digital signal
processing, image and video coding, 3-D video processing and compression,
joint source-channel coding, robust video transmission, distributed source
coding, multiple description coding, and multimedia forensics.
\end{IEEEbiography}

\begin{IEEEbiography}[{\includegraphics[width=1in,height=1.25in,clip,keepaspectratio]{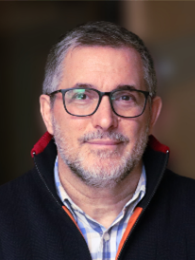}}]{Fernando Pereira} (Fellow, IEEE) graduated in electrical and computer engineering in 1985 and received the M.Sc. and Ph.D. degrees in 1988 and 1991, respectively, from Instituto Superior Técnico, Technical University of Lisbon. He is with the Department of Electrical and Computers Engineering of Instituto Superior Técnico, University of Lisbon, and Instituto de Telecomunicações, Lisbon, Portugal. He is or has been Associate Editor of IEEE Transactions of Circuits and Systems for Video Technology, IEEE Transactions on Image Processing, IEEE Transactions on Multimedia, IEEE Signal Processing Magazine and EURASIP Journal on Image and Video Processing, and Area Editor of the Signal Processing: Image Communication Journal. In 2013-2015, he was the Editor-in-Chief of the IEEE Journal of Selected Topics in Signal Processing. He was an IEEE Distinguished Lecturer in 2005 and elected as an IEEE Fellow in 2008 for “contributions to object-based digital video representation technologies and standards”. He has been elected to serve on the IEEE Signal Processing Society Board of Governors in the capacity of Member-at-Large for 2012 and 2014-2016 terms. He has been IEEE Signal Processing Society Vice-President for Conferences in 2018-2020 and IEEE Signal Processing Society Awards Board Member in 2017. He was the recipient of the 2023 Leo L. Beranek Meritorious Service Award. Since 2013, he is also a EURASIP Fellow for “contributions to digital video representation technologies and standards”. He has been elected to serve on the European Signal Processing Society Board of Directors for a 2015-2018 term. He was the recipient of the 2023 EURASIP Meritorious Service Award. Since 2015, he is also an IET Fellow. He has also held key leadership roles in numerous IEEE Signal Processing Society conferences and workshops, mostly notably serving twice as ICIP Technical Chair in two continents, Hong Kong (2010) and Phoenix (2016). He has been MPEG Requirements Subgroup Chair and is currently JPEG Requirements Subgroup Chair. Recently, he has been one of the key designers of the JPEG Pleno and JPEG AI standardization projects. He has contributed more than 300 papers in international journals, conferences and workshops, and made several tens of invited talks and tutorials at conferences and workshops. His areas of interest are video analysis, representation, coding, description and adaptation, and advanced multimedia services.
\end{IEEEbiography}

\EOD

\end{document}

%% file: glossary.tex
\newacronym{jpeg-pcc}{JPEG PCC}{JPEG Pleno Learning-based Point Cloud Coding Standard}
\newacronym{dl}{DL}{Deep Learning}
\newacronym{sr}{SR}{Super Resolution}
\newacronym{pc}{PC}{Point Cloud}
\newacronym{pcc}{PCC}{Point Cloud Coding}
\newacronym{sqh}{SQH}{Scalable Quality Hyperprior}
\newacronym{esqh}{SRQH}{Scalable Resolution and Quality Hyperprior}
\newacronym{qulpe}{QuLPE}{Quality-conditioned Latent Probability Estimator}
\newacronym{squlpe}{RQuLPE}{Resolution and Quality-conditioned Latents Probability Estimator}
\newacronym{gpcc}{G-PCC}{Geometry-based Point Cloud Compression}
\newacronym{vpcc}{V-PCC}{Video-based Point Cloud Compression}
\newacronym{ae}{AE}{Auto-Encoder}
\newacronym{vae}{VAE}{Variational Auto-Encoder}
\newacronym{vvc}{VVC}{Versatile Video Codec}
\newacronym{sf}{SF}{Scaling Factor}
\newacronym{pct}{PTv2}{Point Transformer v2}
\newacronym{qp}{QP}{quality parameter}
\newacronym{knn}{k-NN}{k Nearest Neighbors}
\newacronym{rd}{RD}{Rate-Distortion}

%% file: 01_intro.tex
\section{Introduction}
\label{sec:intro}
\glspl{pc} have emerged as a fundamental representation for spatial data, comprising collections of points sampled from object surfaces in 3D space. Each point is characterized by its spatial coordinates and may include additional attributes such as color components and surface normals. The prominence of 3D representations, particularly PCs, has grown significantly due to three key factors: their ability to create immersive environments, their support for six-degrees-of-freedom navigation, and their capacity for accurate environmental representation. These characteristics have established PCs as the de facto standard in various applications, including virtual and augmented reality, autonomous navigation, and cultural heritage preservation \cite{camuffo2022recent}.  However, achieving high-fidelity scene representation often requires PCs with millions of points, resulting in substantial storage and bandwidth demands. 

The growing importance of PCs, coupled with their considerable raw data size, has made the development of efficient \gls{pcc} algorithms crucial for practical applications.
\IEEEpubidadjcol 
\gls{pcc} presents unique challenges due to two inherent characteristics of PCs: their unstructured nature and the non-uniform point density resulting from acquisition processes. To address these challenges, standardization efforts have emerged:
\begin{enumerate}
    \item MPEG has developed two distinct standards \cite{graziosi2020overview}:
    \begin{itemize}
        \item Geometry-based Point Cloud Compression (G-PCC), originally targeting static or dynamically acquired PCs (e.g. LiDAR PCs);
        \item Video codec-based Point Cloud Compression (V-PCC), originally targeting dynamic PCs.
    \end{itemize}
    \item JPEG is finalizing the development of \gls{jpeg-pcc} \cite{guarda2024jpeg}, the first learning-based standard for static \gls{pcc}.
\end{enumerate}

JPEG's effort in particular, being \gls{jpeg-pcc} learning-based, aims to create a standard that delivers competitive \gls{rd} performance while providing effective compressed domain representations for both human visualization and machine processing \cite{seleem2023deep}.

While RD performance is crucial, real-world applications demand additional features, including stream scalability. Scalability enables serving content at various qualities, resolutions, or framerates through a single bitstream, rather than maintaining multiple independent streams. This capability becomes particularly valuable given the heterogeneity of receiving devices, which often operate under different network conditions and hardware constraints. Implementing this feature can thus allow a codec to properly adapt to a wide variety of receiving conditions.
A scalable bitstream is generally organized into a base layer providing minimal rate decoding capabilities, and multiple enhancement layers allowing progressive refinement of the visual fidelity.

The growing importance of scalability in point cloud coding has prompted standardization organizations to incorporate this feature into their codec requirements \cite{jpeg-pleno-cttc} and to propose extensions to existing solutions \cite{vpccscal}. In particular, according to these requirements, some of the most relevant forms of scalability for static PCs are:
\begin{itemize}
    \item Quality Scalability: Enhancement layers improve visual fidelity without affecting resolution.
    \item Resolution Scalability: Enhancement layers increase content resolution, with quality improvement being a secondary effect.
\end{itemize}

Recent advancements in this direction can be found in \cite{mari2024point} 
where the authors presented a method for geometry quality scalability in \gls{jpeg-pcc} by working in the latent domain. Even if only quality scalability is considered, the approach demonstrates how visual fidelity can be progressively enhanced with minimal additional information by correctly exploiting the previously encoded base and enhancement layers.

Some PC codecs (e.g. \gls{jpeg-pcc}) reduce the resolution of the input PC as a form of point domain quantization, controlled by a scaling factor $sf$. Downscaling is thus an effective RD tuning strategy for PCs since it helps obtaining lower bitrates through the reduction of the information in the content. Additionally, it can help increasing the spatial redundancy in sparser PCs (making them more efficient to compress). Therefore, reducing the resolution of the input is not only useful for addressing the needs of devices that don't support higher resolutions, but it is also an effective way to improve RD performance and increase the number of covered rate points. 

While \gls{jpeg-pcc} includes a learned super-resolution module to reverse this process, it does not provide true resolution scalability as it operates without enhancement layers, potentially resulting in lower visual fidelity compared to non-super-resolved reconstructions.
Additionally, existing solutions, like Mari et al. \cite{mari2024point} only address quality scalability, so they cannot handle bitstreams for PC coded using different scaling factors. This limitation restricts the codec's adaptability and prevents full utilization of its capabilities when scalable bitstreams are required. 

To address these challenges, this paper proposes Scalable Resolution and Quality Hyperprior (\gls{esqh}), a novel approach providing joint resolution and quality scalability for PC geometry coding.  Integrated into  
\gls{jpeg-pcc}, \gls{esqh} enables scalability while maintaining compatibility with the base codec's non-scalable operation mode. This integration is achieved by replacing the hyper-analysis and hyper-synthesis transforms with the \gls{squlpe} model during enhancement layer coding.

The main key innovations and contributions brought by \gls{esqh} are:
\begin{enumerate}
    \item Joint scalability of point cloud resolution and quality: \gls{esqh} enables \gls{jpeg-pcc} to serve diverse devices through a single scalable bitstream. 
    \item Minimal RD performance impact: the proposed approach pays a small price for scalability compared to non-scalable \gls{jpeg-pcc} where, for one RD point, the PC can be decoded only at one specific resolution and quality.
    \item Reduced memory requirements: \gls{esqh} requires smaller neural networks compared to state-of-the-art solutions.
    \item Minimal computational overhead: when using \gls{esqh} encoding time increases by less than $<$10\% and decoding time by less than $<$20\% per decoded enhancement layer w.r.t. \gls{jpeg-pcc}.
    \item Generic design: \gls{esqh} can be easily integrated in autoencoder-based deep learning codecs for other multimedia modalities.
    \item Correlated latent spaces: this work shows that the correlation between the latent spaces in the different \gls{jpeg-pcc} models arises from sequential training. This property helps providing effective scalability algorithms and it can be easily introduced in other autoencoder based codecs such as \cite{balle2018variational, minnen2018joint, quach2020improved}
\end{enumerate}

The rest of the paper is organized as follows. Section~\ref{sec:related} overviews
the current state-of-the-art in \gls{pc} coding, while Section~\ref{sec:vm} describes the \gls{jpeg-pcc} codec, serving as reference and base layer codec for the proposed approach and the \gls{sqh} algorithm proposed in \cite{mari2024point} which is improved upon in this work. 
Section~\ref{sec:extending_sqh} proposes the novel \gls{esqh} algorithm for joint resolution and quality scalability across a wide variety of coding configurations, and the newly proposed \gls{squlpe} which is the main neural network used in \gls{esqh}. Finally, Section~\ref{sec:rqulpe results} reports
and analyses the most relevant experimental results and Section~\ref{sec:conclusions}
discusses future work directions.

%% file: 02_previous_works.tex
\section{Related Works}
\label{sec:related}
State-of-the-art PCC encompasses various approaches, ranging from traditional signal processing techniques to modern learning-based solutions. Among the conventional approaches, the most relevant are G-PCC and V-PCC \cite{graziosi2020overview}, the two MPEG standards for \gls{pcc}.
G-PCC, or Geometry-based Point Cloud Compression, leverages octree representations for efficient geometry coding, and uses predictive or hierarchical transforms for attribute coding. On the other hand, V-PCC, or Video-based Point Cloud Compression, projects 3D PC data into the 2D domain, creating images that represent the geometry and texture information, which are compressed using very efficient and established video codecs like HEVC and VVC \cite{bross2021overview}.
While G-PCC inherently supports resolution scalability for both geometry and attributes, achieving scalability in V-PCC presents challenges due to its reliance on video coding frameworks. Although MPEG has initiated investigations into various scalability techniques for V-PCC \cite{vpccscal}, these features remain to be specified in the current version of the standard.

More recently, the advent of \gls{dl} has revolutionized PC compression, yielding numerous high-performing solutions \cite{guarda2024jpeg,quach2020improved,wang2021lossy, wang2022sparse, liu2022pcgformer}. Nevertheless, among the many \gls{dl}-based PCC solutions, few approaches address any form of scalability. DL-PCSC \cite{guarda2020point} implements quality scalability by channelwise partitioning of latent representations, enabling progressive quality enhancement through incremental transmission. However, this approach faces limitations: the requirement for zero-padding untransmitted latents constrains the latent space design, and the reduced latent space dimensionality at lower rates leads to reduced modeling capabilities \cite{balle2018variational}. These constraints significantly impact the rate-distortion performance making scalability less appealing.

GRASP-Net \cite{pang2022grasp} offers an alternative approach, by implementing a \gls{dl}-based enhancement layer atop a G-PCC base layer. However, scalability is limited to this two-layer structure, and the extremely low-resolution base layer may prove impractical for real-world applications.

The work by Ulhaq et al. \cite{ulhaq2024scalable} implements scalability by adapting content for human and machine consumption. In particular, the base layer can be used to solve computer vision tasks (e.g. PC classification) while the enhancement layer allows reconstructing the PC for human visualization. This approach thus addresses scalability requirements that are different from those explored in this work.

SparsePCGC \cite{wang2022sparse} and its successor, Unicorn \cite{wang2024versatile1,wang2024versatile2}, represent significant advances in resolution scalability for PCC, due to their inherently multiscale nature. At the encoder side, Unicorn employs a hierarchical downscaling approach, encoding the information necessary for losslessly upscaling it at the decoder.
When this enhancement information is unavailable, the decoder employs a lossy thresholding strategy for upscaling.
By dividing the bitstream in different enhancement layers, required to upscale the PC, Unicorn achieves resolution scalability. Furthermore, Unicorn has a very competitive coding performance when compared with other PC codecs. The intrinsic scalability mechanism of Unicorn, which results from its architecture, contrasts with \gls{esqh}, which offers a modular, plug-and-play solution applicable to various codecs.

Additionally, it is also important to mention that recently the MPEG group issued the call for proposal for the new AI-based PCC \cite{aigccfp} which is currently under development.

%% file: 03_vm.tex
\section{JPEG Pleno Point Cloud Geometry Codec and Previous Extension for Quality Scalability}
\label{sec:vm}

\begin{figure*}
    \centering
    \includegraphics[width=.9\textwidth]{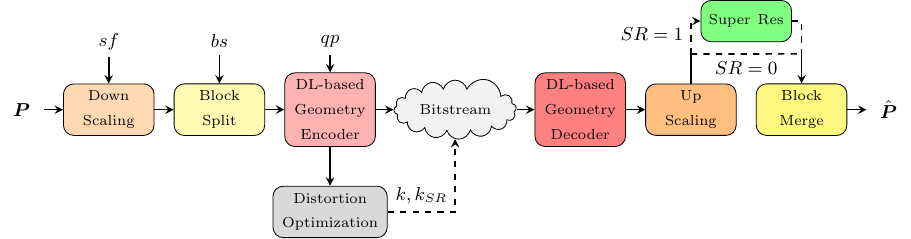}
    \caption{High-level scheme of the coding procedure for PC geometry in \gls{jpeg-pcc}.}
    \label{fig:high-level-scheme}
\end{figure*}

\label{sec:jpegpleno}
The \gls{esqh} method proposed in this paper is implemented on top of the verification model software for the \gls{jpeg-pcc} standard \cite{jpeg-pleno}, which will be presented next in this section. After that, \gls{sqh} \cite{mari2024point}, a previously proposed solution for implementing quality scalability in \gls{jpeg-pcc} for geometry coding, which served as the basis for the \gls{esqh} method, will be described.

\subsection{JPEG Pleno Point Cloud Codec}

\gls{jpeg-pcc} is the JPEG standard for PC coding, which uses a learning-based approach for coding both PC geometry and color attributes \cite{guarda2023point}.
The geometry coding in \gls{jpeg-pcc} utilizes a deep learning model structured as an autoencoder, complemented with a variational autoencoder model that determines a mean and scale hyperprior that improves the performance for entropy coding the compressed domain latent representation \cite{minnen2018joint}. To enhance compression performance, particularly for sparse PCs and lower-rate coding scenarios, \gls{jpeg-pcc} incorporates two additional tools:
\begin{enumerate}
    \item A down-scaling module using a scaling factor parameter, $sf$.
    \item A deep learning-based super-resolution (SR) module to improve reconstruction quality when down-scaling is applied.
\end{enumerate}

\gls{jpeg-pcc} adopts a sparse tensor representation \cite{choy20194d} for geometry coding, offering advantages in both computational complexity and rate-distortion performance. In this representation, PCs are described as a tuple $\bm{x}=(\bm{x}_C, \bm{x}_F)$, where $\bm{x}_C$ represents the coordinates of non-empty voxels, and $\bm{x}_{F}$ denotes the corresponding features (initially set to "1" to indicate occupied voxels).

For encoding the color attributes, JPEG
PCC projects texture patches onto an image (similarly to V-PCC \cite{V-PCC}) which is then coded using the emerging JPEG AI codec \cite{jpeg-ai}.
Since the focus of this paper lies in geometry coding, the remaining of this section will focus only on this component.

A high-level description of the full geometry coding and decoding procedures is shown in Fig.~\ref{fig:high-level-scheme}. Specifically, to encode the geometry of a point cloud $\bm{P} \in \mathbb{R}^3$, the encoder performs the following operations:

\begin{figure*}
    \centering
    \includegraphics[width=.9\textwidth]{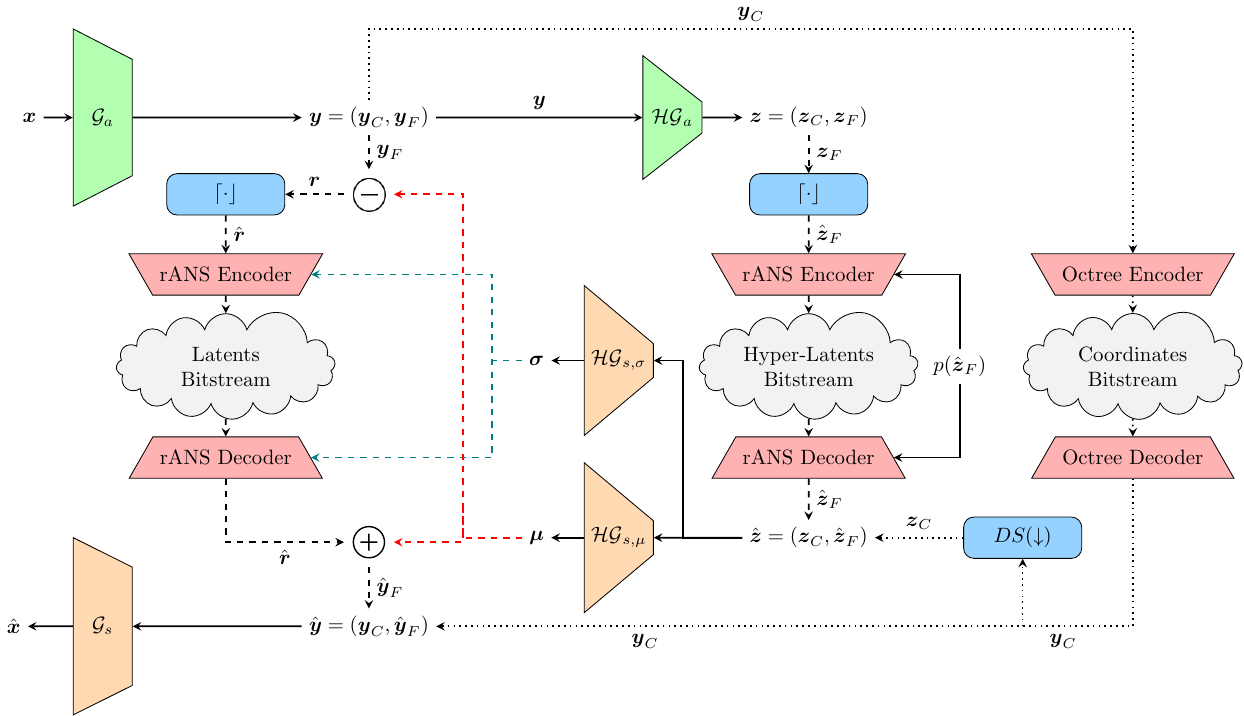}
    \caption{Model architecture of the deep learning based codec in \gls{jpeg-pcc} (\gls{dl}-based Geometry Encoder and \gls{dl}-based Geometry Decoder in Fig.~\ref{fig:high-level-scheme}).}
    \label{fig:dl-scheme}
\end{figure*}

\begin{enumerate}[label=E\arabic*.]
    \item \textit{Downscaling}: The input PC is downscaled by a factor $sf$ through the operation $\boldsymbol{P}^\prime = \lceil \boldsymbol{P}/sf \rfloor$.
    \item \textit{Block Split}: The downscaled points are divided into non-overlapping blocks $\x_{l, C} \in \mathbb{R}^3, l \in {1, \dots, N}$ of size $bs$, such that $\boldsymbol{P}^\prime = \bigcup_{l=1}^{N} \x_{l, C}$.
    \item \textit{Sparse Tensor Construction}: For each block, a sparse tensor representation $\x_l = (\x_{l, C}, \x_{l, F})$ is created, where $\x_{l, F}$ contains ones to indicate occupied voxels.
    \item \textit{DL-Based Encoding}: The blocks are processed through the deep learning-based coding procedure to generate the bitstream
    \item \textit{Distortion Optimization}: Two parameters per block, $k_l$ and $k_{SR, l}$, are computed and added to the bitstream. These parameters represent the optimal number of points to be retained in the decoded block (with and without super-resolution) to minimize a chosen distortion metric.
\end{enumerate}

At the decoder side, the PC reconstruction process consists of the following operations:
\begin{enumerate}[label=D\arabic*.]
    \item \textit{DL-Based Decoding}: The decoder reconstructs the blocks $\hat{\x}_l$ by inputting the compressed domain latent representation, extracted from the bitstream, in the DL-based decoder.
    \item \textit{Top-K Points Selection}: For each decoded block $\hat{\x}_l$, only the $k_l$ points with the highest occupancy probabilities are retained, ensuring optimal point selection.
    \item \textit{Upscaling}: The blocks are upscaled according to scaling factor $sf$ (included in the bitstream) to restore the original spatial resolution.
    \item \textit{Super-Resolution}: When super-resolution is enabled ($SR=1$), the upscaled blocks are processed through the SR network to obtain enhanced blocks $\hat{\x}_{SR, l}$.
    \item \textit{Post-SR Top-K Point Selection}: From each super-resolved block, the $k_{SR,l}$ points with the highest probability values are selected, ensuring optimal point selection.
    \item \textit{Block Merge}: Finally, all processed blocks are merged to reconstruct the complete point cloud geometry $\hat{\boldsymbol{P}}$.
\end{enumerate}

The deep learning-based encoding (and decoding) process for each block $x_l$, is illustrated in Fig.~\ref{fig:dl-scheme}. It consists of the following sequence of operations:

\begin{enumerate}[label=E\arabic*.]
    \item \textit{Latents Generation}: Generate latents $\y_l = (\y_{l, C}, \y_{l, F})$ through the analysis transform $\mathcal{G}_a$, expressed as $\y_l=\mathcal{G}_a(\x_l)$.
    \item \textit{Coordinates Encoding}: code the latent coordinates $\y_{l, C}$ using an octree encoder to generate the coordinates bitstream.
    \item \textit{Hyper-Latents Generation}: Generate hyper-latents $\z_l$ using the hyper-analysis transform $\mathcal{HG}_a$, where $\z_l=\mathcal{HG}_a(\y_l)$.
    \item \textit{Hyper-Latents Quantization}: Quantize the hyper-latent features to obtain $\hat{\z}_{l, F} = \lceil \z_{l, F} \rfloor$.
    \item \textit{Entropy Coding}: Apply rANS entropy coding to the hyper-latents according to a fully factorized prior $p(\hat{\z}_{l, F})$ to generate the hyper-latents bitstream.
    \item \textit{Sparse Tensor Construction}: Reconstruct the quantized hyper-latents' sparse representation as $\hat{\z}_l = (\z_{l, C}, \hat{\z}_{l, F})$.
    \item \textit{Latents Distribution Estimation}: Process $\hat{\z}_l$ through hyper-synthesis transforms $\mathcal{HG}_{s, \mu}$ and $\mathcal{HG}_{s, \sigma}$ to estimate Gaussian parameters $\boldsymbol{\mu}_l = \mathcal{HG}_{s, \mu}(\hat{\z}_l), \boldsymbol{\sigma }_l= \mathcal{HG}_{s, \sigma}(\hat{\z}_l)$. 
    \item \textit{Residual Encoding}: Calculate and encode quantized residuals $\boldsymbol{r}_l= \lceil \y_{l,F} - \boldsymbol{\mu}_l \rfloor$ using $\mathcal{N}(\boldsymbol{0}, \boldsymbol{\sigma}_l)$ to produce the final latents bitstream.
\end{enumerate}

Conversely, a receiver that needs to decode the blocks from the bitstream will have to:

\begin{enumerate}[label=D\arabic*.]
    \item \textit{Coordinates Decoding}: Losslessly decode $\y_{l, C}$ from the coordinates bitstream.
    \item \textit{Hyper-latents Decoding}: Entropy decode $\hat{\z}_{l, F}$ from the hyper-latents bitstream using the probability distribution $p(\hat{\z}_{l, F})$.
    \item \textit{Coordinates Down-scaling}: Down-scale $\y_{l, C}$ by a factor of 4 (as determined by the stride parameters in $\mathcal{HG}_a$'s convolutional layers) to obtain $\z_{l, C}$.
    \item \textit{Hyper-Latents Sparse Tensor Construction}:  Build the sparse representation of hyper-latents as $\hat{\z}_l = (\z_{l, C}, \hat{\z}_{l, F})$.
    \item \textit{Latents Distribution Estimation}:  Compute Gaussian parameters using hyper-synthesis transforms as $\boldsymbol{\mu}_l = \mathcal{HG}_{s, \mu}(\hat{\z}_l), \boldsymbol{\sigma }_l= \mathcal{HG}_{s, \sigma}(\hat{\z}_l)$.
    \item \textit{Residuals Decoding}: Entropy decode $\boldsymbol{r}_l$ from the latents' bitstream using $\mathcal{N}(\boldsymbol{0}, \boldsymbol{\sigma}_l)$.
    \item \textit{Latent Features Reconstruction}: Recover the latent features $\hat{\y}_{l, F} = \boldsymbol{\mu}_l + \boldsymbol{r}_l$.
    \item \textit{Latents Sparse Tensor Construction}: Reconstruct the sparse representation of latents as $\hat{\y}_l = (\y_{l, C}, \hat{\y}_{l, F})$.
    \item \textit{Block Reconstruction}: Apply the synthesis transform $\mathcal{G}_s$ to the decoded latents to determine the probability for the occupancy state of each voxel in the reconstruct the block: $\hat{\x}_l = \mathcal{G}_s(\hat{\y}_l)$. 
\end{enumerate}

The model training follows an end-to-end approach incorporating all previously described operations except for two differences: quantization is replaced by a differentiable approximation and entropy coding is removed, to ensure full model differentiability. The training utilizes a rate-distortion optimization framework defined by the loss function:
\begin{equation}
    \mathcal{L}(\bm{x}, \hat{\bm{x}}, \bm{y}, \bm{z}) = \mathcal{D}(\bm{x}, \hat{\bm{x}}) + \lambda\mathcal{H}(\bm{y}, \bm{z}),
\end{equation}
where $\mathcal{D}(\cdot, \cdot)$ is the distortion, measured as the focal loss \cite{lin2017focal}, $\mathcal{H}(\cdot, \cdot)$ denotes the entropy of the bitstream components under the probability distributions $p(\hat{\bm{z}})$ and $p(\bm{y}|\hat{\bm{z}})$, and $\lambda$ controls the rate-distortion trade-off.
Generally, one model is trained for each RD point corresponding to one value of $\lambda$. In \gls{jpeg-pcc}, five different coding models are trained to support the defined range of tradeoffs.
The training procedure is carried out by sequentially spanning the chosen values of $\lambda \in \{0.0025, 0.005, 0.01, 0.025, 0.05\}$, using the checkpoint for the previous $\lambda$ as a starting point, progressively moving from the lowest value (highest rate/quality) to the highest one (lowest rate/quality). These five models naturally define a quality parameter $qp \in \{1, \dots , 5\}$, with $qp = 1$ corresponding to $\lambda = 0.05$ (lowest rate/quality) and $qp = 5$ to $\lambda = 0.0025$ (highest rate/quality).

\subsection{Scalable Quality Hyperprior}

The \gls{esqh} method proposed in this paper follows a previous work \cite{mari2024point} that introduced a quality scalability algorithm, known as \gls{sqh}. \gls{sqh} constructs a quality scalable bitstream by leveraging information from latents $\y_i$ obtained at a lower \gls{qp} ($qp=i$) to predict probability distributions for latents $\y_j$ at a higher \gls{qp} ($qp=j$).

Starting from a low-quality base layer of latents $\y_i$, which have already been encoded, the encoder must execute the following sequence of steps to generate a new enhancement layer:

\begin{enumerate}[label=E\arabic*.]
    \item \textit{Higher Quality Latents Generation}: Generate new latents $\y_{j}$ using the \gls{jpeg-pcc} coding model with $qp = j > i$.
    \item \textit{Latents Distribution Estimation}: Predict the means and standard deviations of the
latents $\y_{j}$ based on the previous latents $\y_{i}$, using the \gls{dl}-based \gls{qulpe} model
(detailed in \cite{mari2024point}) as $\boldsymbol{\mu}_j$, $\boldsymbol{\sigma}_j = QuLPE(\hat{\y}_i, i, j)$, under the assumption of independently distributed Gaussian latents, $P(\y_j|\hat{\y}_i)$.
    \item \textit{Entropy Coding}: Generate the \gls{sqh} bitstream by entropy encoding $\y_j$ using $\boldsymbol{\mu}_j$, $\boldsymbol{\sigma}_j$.
\end{enumerate}

\gls{sqh} employs a Mean and Scale Hyperprior entropy model, analogous to \gls{jpeg-pcc}. The key distinction lies in \gls{sqh}'s utilization of previously decoded latents $\hat{\y}_i$ as side information, rather than hyper-latents $\hat{\z}_j$.

To reconstruct the higher rate/quality PC, the decoder, which can access the base layer information $\hat{\y}_i$, performs the following decoding procedure:
\begin{enumerate}[label=D\arabic*.]
    \item \textit{Latents Distribution Estimation}: Derive $\boldsymbol{\mu}_{j}$, $\boldsymbol{\sigma}_{j} = QuLPE(\hat{\y}_i, i, j)$ from the base layer information $\hat{\y}_i$ using the \gls{qulpe} model.
    \item \textit{Higher Quality Latents Decoding}: Decode the higher quality latents $\hat{\y}_{j}$ by applying a rANS decoder to the \gls{sqh} bitstream using the estimated distribution.
    \item \textit{Higher Quality PC Reconstruction}: Reconstruct the higher quality PC by processing the decoded latents through the \gls{jpeg-pcc} synthesis transform as $\hat{\x}_{j} = \mathcal{G}_{s, j}(\hat{\y}_{j})$.
    \item \textit{Super Resolution}: If specified in the coding parameters the Super Resolution model is used to enhance the decoded blocks
\end{enumerate}

While \gls{sqh} effectively handles quality scalability through latents refinement, it faces limitations when dealing with \gls{jpeg-pcc}'s downscaling strategy. The challenge arises because varying $sf$ produces latents at different resolutions, a scenario not supported by \gls{sqh}'s U-Net-based \gls{qulpe} model, which requires consistent input-output dimensions. This architectural constraint, coupled with the absence of a multi-resolution handling strategy, restricts \gls{sqh}'s practical applicability. 

The next sections introduce and evaluate \gls{esqh}, an enhanced framework that addresses these limitations by enabling joint quality and resolution scalability in the latent domain. These functional advantages are relevant in the framework of the \gls{jpeg-pcc} codec, but also for the generalization of the \gls{esqh} method for other autoencoder-based codecs.

%% file: 04_method.tex
\begin{figure*}[!t]
    \centering
    \begin{minipage}[c]{.28\textwidth}
    \centering
    \includegraphics[height=\columnwidth]{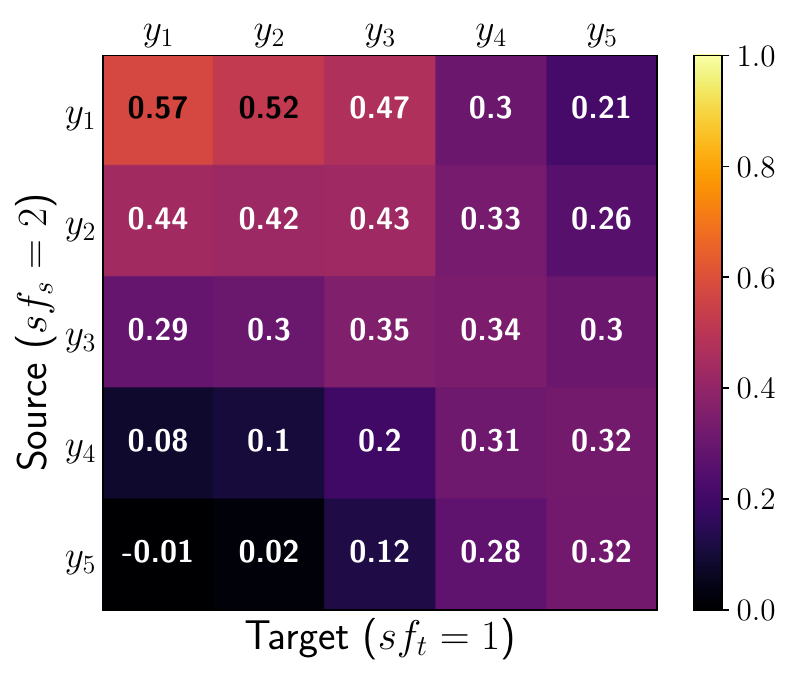}
    \subcaption{Sequentially trained ($sf_s \neq sf_t$).}
    \label{fig:corr_matrix_sf_seq}
    \end{minipage}    
    \hspace{1cm}
    \begin{minipage}[c]{.28\textwidth}
    \centering
    \includegraphics[height=\columnwidth]{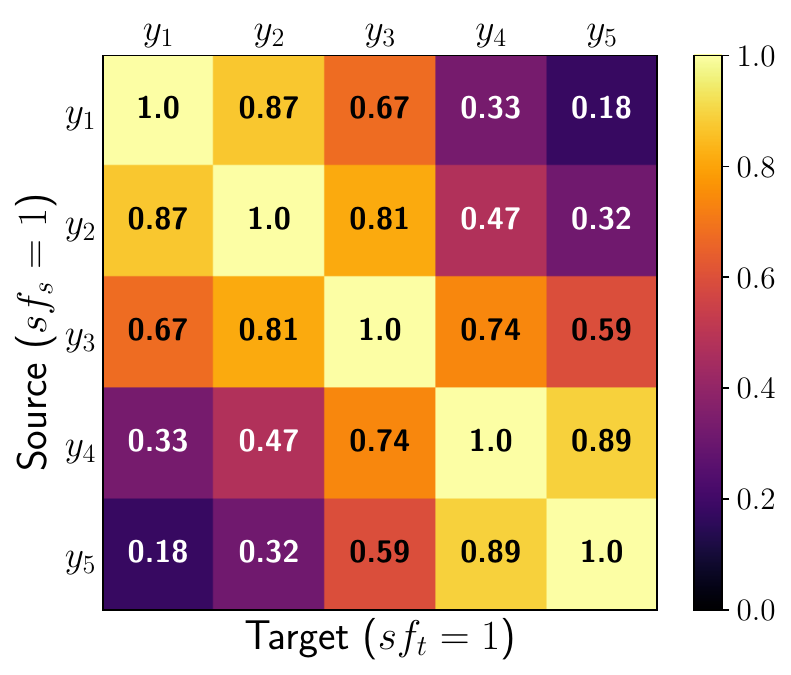}
    \subcaption{Sequentially trained ($sf_s = sf_t$).}
    \label{fig:corr_matrix_q_seq}
    \end{minipage}    
    \hspace{1cm}
    \begin{minipage}[c]{.28\textwidth}
    \centering
    \includegraphics[height=\columnwidth]{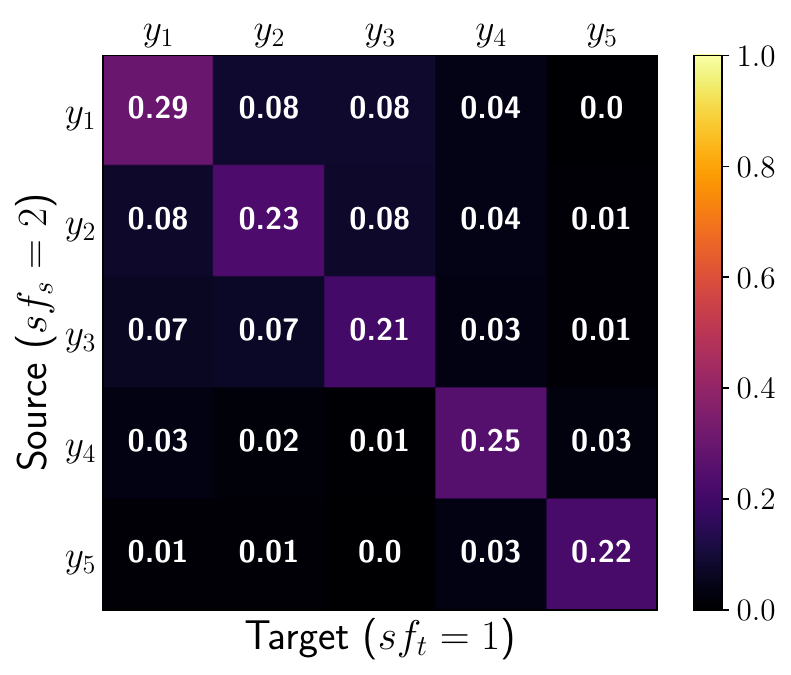}
    \subcaption{Independently trained.}
    \label{fig:corr_matrix_sf_ind}
    \end{minipage}
    \label{fig:corr_matrix_sf}
    \caption{Average cosine similarity between corresponding latents produced by the five different \gls{jpeg-pcc} coding models.}
\end{figure*}

\section{Scalable Resolution and Quality Hyperprior}
\label{sec:extending_sqh}
Previous research \cite{mari2024point} revealed a significant correlation between latents encoded with varying $qp$ parameters. This fundamental property is crucial to guarantee \gls{sqh}'s effectiveness as it enables a single model to manage diverse coding parameter combinations. However, this raises an important question: does this correlation persist when latents exhibit different resolutions due to varying scaling factors, potentially alongside quality differences? 
To rigorously investigate this relationship, an analysis was conducted using cosine similarity measurements between latents encoded across different combinations of quantization parameters $qp$ and scaling factors $sf$. For clarity in subsequent discussions, a simplified notation convention is adopted: lower-rate latents and their associated parameters are designated as "source" elements, denoted with the suffix $s$ (e.g., $\y_s$), while higher-rate latents and their corresponding parameters are termed "target" elements, indicated by the suffix $t$ (e.g., $\y_t$).

\subsection{Correlation between the latents}

Given blocks $\x$ in the validation dataset, latent vectors
$\boldsymbol{y}_{i} = Enc(\x, sf_s, i)$ and $\boldsymbol{y}_{j} = Enc(\x, sf_t, j)$
are generated, where $sf_s, sf_t$ denote the scaling factors,
$i, j$ represent the quality parameters, and $Enc(\cdot, \cdot, \cdot)$
represents the encoding function that performs down-scaling and applies the analysis transform to the input block.
Following the methodology established for \gls{sqh}, cosine similarities between different compressed representations of identical blocks were evaluated across varying quality parameters and resolutions.

The analysis considers source latents encoded with parameters $qp_s=i, sf_s$ and target latents with $qp_t=j, sf_t$. With $\y_s = Enc(\x, sf_s, i)$ and $\y_t = Enc(\x, sf_t, j)$, the matrix coefficients at position $i, j$ were derived as the average cosine similarity between latents across all the blocks of the validation dataset. For cases where $sf_s \neq sf_t$, the source and target coordinates do not match due to the resolution difference. In such instances, the lower resolution coordinates were upscaled and the cosine similarities were computed relative to nearest neighbors within a radius of 2, selected based on the ratio between $sf_s$ and $sf_t$.

This analysis generated three distinct similarity matrices. The first two matrices (shown in Figs.~\ref{fig:corr_matrix_sf_seq},~\ref{fig:corr_matrix_q_seq}) were computed using latents from \gls{jpeg-pcc}v4.0, featuring sequentially trained models. Fig.~\ref{fig:corr_matrix_sf_seq} corresponds to $sf_s=2, sf_t=1$, while Fig.~\ref{fig:corr_matrix_q_seq} represents $sf_s=sf_t=1$. The third matrix (shown in Fig.~\ref{fig:corr_matrix_sf_ind}) utilizes latents from independently trained models, maintaining identical architecture, training data, and parameters as the sequential case. This comparison serves to determine whether latent alignment emerges as a consequence of sequential training.

The results in Fig.~\ref{fig:corr_matrix_sf_seq}, representing the sequential case with $sf_s\neq sf_t$, demonstrate positive cosine similarity between latents, although lower than cases with identical scaling factors (Fig.~\ref{fig:corr_matrix_q_seq}). This effect becomes particularly pronounced with increasing disparities between $qp_s$ and $qp_t$.
Notably, configurations where $qp_t < qp_s$ exhibit lower cosine similarity compared to cases where $qp_t > qp_s$, indicating suboptimal conditions for \gls{esqh} operation under such parameters configurations.

The analysis of Fig.~\ref{fig:corr_matrix_sf_ind} reveals that independent model training results in completely unaligned latent spaces, thereby demonstrating that sequential training is the key mechanism enabling latent space alignment. This alignment property facilitates the mapping between latent domains generated under different coding configurations, which is an important factor for the effectiveness of the proposed \gls{esqh} algorithm, showing the advantage of using sequential training.

\subsection{Design of the Scalable Resolution and Quality Hyperprior}

Designing \gls{esqh} to handle latents at different resolutions requires effectively encoding $\y_t$ given known source latents $\hat{\y}_s$.
Since these latents are sparse tensors such that $\y_{s, F} \neq \y_{t, F}$, $\y_{s, C} \neq \y_{t, C}$ then \gls{esqh} requires two main modules.
\begin{enumerate}
    \item A coordinates coding module (referred to as \gls{squlpe}-C) capable of encoding target coordinates $\y_{t, C}$ at higher resolutions w.r.t. the source coordinates $\y_{s, C}$.
    \item A features coding module (referred to as \gls{squlpe}-F) that can estimate probability distributions for target latent representations $\y_{t, F}$ based on the source latents $\y_s$.
\end{enumerate}    

The coordinates coding module becomes essential when $sf_s > sf_t$, as this condition results in a higher resolution for the target block compared to the source block. This resolution disparity is mirrored in the resolution of the latent representations, necessitating the encoding of supplementary information to enable the decoder to reconstruct the higher-resolution coordinates accurately. For this purpose, a lossless geometry coding solution was adopted.
In particular, a set of plausible high-resolution coordinates $\boldsymbol{\Tilde{y}}_{t, C}$ are obtained from $\y_{s, C}$ and then the ground truth target occupancy $GT(\boldsymbol{\Tilde{\y}}_{t, C})$ is losslessly encoded using a probability distribution predicted by the \gls{squlpe}-C model.

On the other hand, \gls{squlpe}-F is needed because the values of the source and target latents are different in both resolution and quality scalability scenarios. This means that a network that can model the target values given the source latents is required. 

\begin{figure*}
    \includegraphics[width=\textwidth]{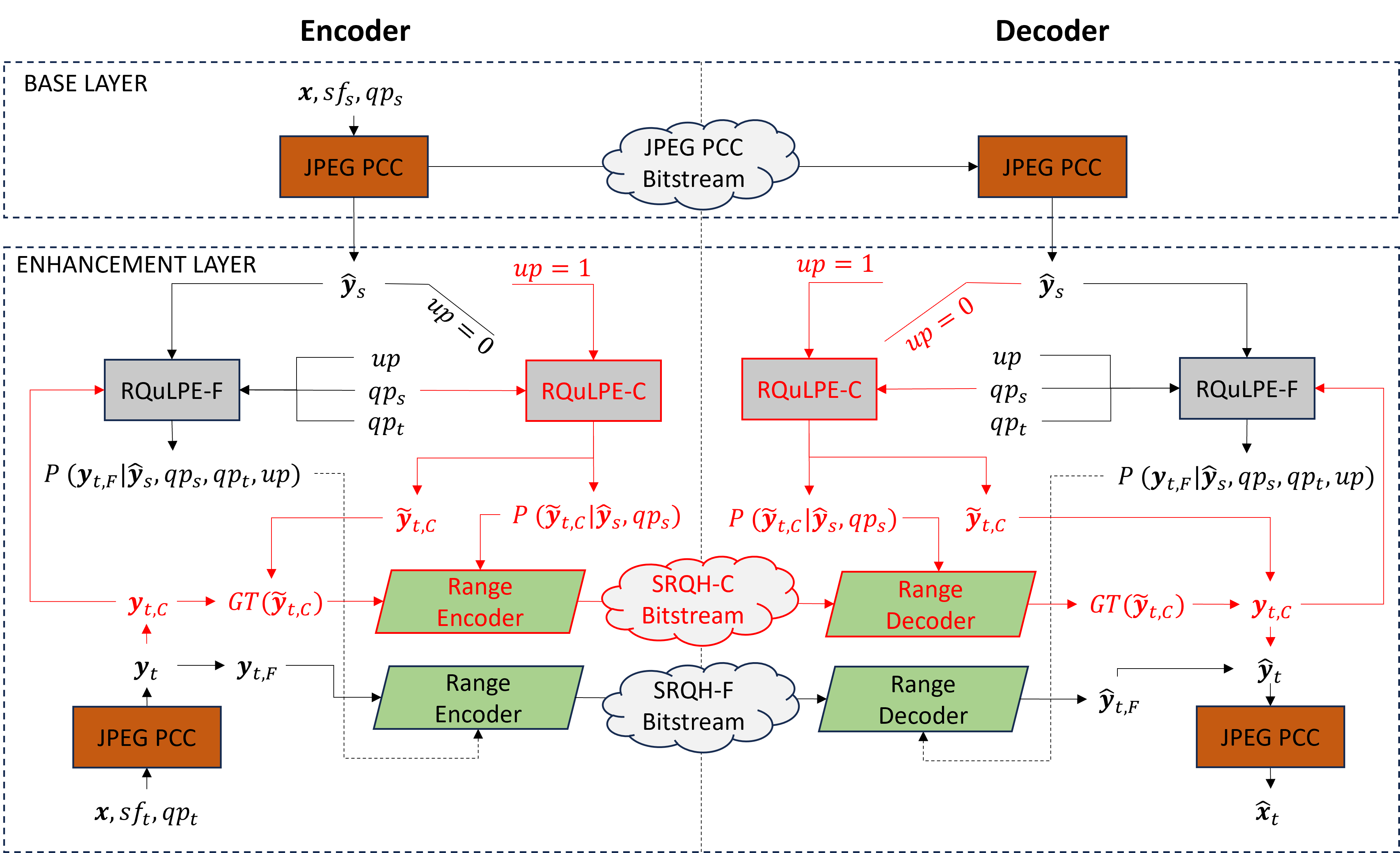}
    \caption{Scalable Resolution and Quality Hyperprior coding scheme. In red are the blocks that are relative to the encoding of the coordinates.}
    \label{fig:esqh}
\end{figure*}
The described components form the foundation of \gls{esqh}, a generalization of \gls{sqh}, that extends its functionality to handle varying latents' resolutions. This enhanced coding scheme replaces the original \gls{qulpe} model with \gls{squlpe}, which comprises the two specialized components: \gls{squlpe}-C and \gls{squlpe}-F.

As illustrated in Fig.~\ref{fig:esqh}, the algorithm begins with a base layer generated using a low-rate \gls{jpeg-pcc} bitstream, containing latents, hyper-latents, and coordinates bitstreams (detailed in Section~\ref{sec:jpegpleno}). Enhancement layers are then constructed by stacking \gls{esqh} bitstreams, enabling progressive decoding at higher resolutions and/or qualities. 

When $sf_s = sf_t$, the \gls{esqh} bitstream is equivalent to the standard \gls{sqh} latents bitstream. However, when $sf_s\neq sf_t$, an additional bitstream for upsampling the latents' coordinates is required. Additionally, the receiver does not need a side bitstream for decoding hyper-latents $\hat{\z}$ since $\hat{\y}_s$ serves as the new side information.
More specifically, given a base layer $\hat{\y}_{s}$ encoded with parameters $qp_s, sf_s$ using the \gls{jpeg-pcc} coding procedure, the encoder executes the following sequence to generate higher rate/quality layer bitstreams:

\begin{enumerate}[label=E\arabic*.]
    \item \textit{Higher Rate Latents Generation}: Encode the PC $\x$ using \gls{jpeg-pcc} with target parameters $sf_t$ and $qp_t$ to obtain $\y_{t}$.
    \item \textit{Upsampled Coordinates Probability Estimation}: For cases where $sf_s \neq sf_t$, compute candidate coordinates $\boldsymbol{\Tilde{y}}_{t, C}$ from $\hat{\y}_{s, C}$ and estimate $P(\boldsymbol{\Tilde{y}}_{t, C}|\hat{\y}_{s}, qp_s)$ using \gls{squlpe}-C.
    \item \textit{Coordinates Entropy Encoding}: Generate the \gls{esqh}-C bitstream by entropy encoding the ground truth occupancy $GT(\boldsymbol{\Tilde{\y}}_{t, C})$ using the estimated probability distribution $P(\boldsymbol{\Tilde{y}}_{t, C}|\hat{\y}_{s}, qp_s)$.
    \item \textit{Latents Features Probability Estimation}: Use \gls{squlpe}-F to estimate $\boldsymbol{\mu}_{t}, \boldsymbol{\sigma}_{t}$, assuming independently distributed Gaussian latents: $P(\y_{t, F}|\hat{\y}_{s}, qp_s, qp_t, up) = \mathcal{N}(\boldsymbol{\mu}_{t}, \boldsymbol{\sigma}_{t})$ where $up$ is a variable that specifies if latents super-resolution is required.
    \item \textit{Latents Features Entropy Encoding}: Generate the \gls{esqh}-F bitstream by entropy encoding $\y_{t, F}$ using the estimated parameters $\boldsymbol{\mu}_{t}, \boldsymbol{\sigma}_{t}$.
\end{enumerate}

The entropy modeling procedure in \gls{esqh} closely resembles that of \gls{jpeg-pcc}, as both approaches utilize a hyperprior to estimate a Gaussian prior for the latents. The key distinction in \gls{esqh} is the use of $\hat{\y}_{s}$ as the hyperprior, rather than the hyper-latents $\hat{\boldsymbol{z}}_{t}$ employed in \gls{jpeg-pcc}.

The decoder, after decoding $\hat{\y}_{s}$, can obtain $\hat{\y}_{t}$ through the following sequence of steps:

\begin{enumerate}[label=D\arabic*.]
    \item \textit{Upsampled Coordinates Probability Estimation}: If $sf_s \neq sf_t$, compute the candidate coordinates $\boldsymbol{\Tilde{y}}_{t, C}$ from $\hat{\y}_{s, C}$ and estimate $P(\boldsymbol{\Tilde{y}}_{t, C}|\hat{\y}_{s}, qp_s)$ using \gls{squlpe}-C.
    \item \textit{Coordinates Entropy Decoding}: Entropy decode the ground truth $GT(\boldsymbol{\Tilde{y}}_{t, C})$ from the \gls{esqh}-C bitstream using the estimated probability distribution $P(\boldsymbol{\Tilde{y}}_{t, C}|\hat{\y}_{s}, qp_s)$ 
    \item \textit{Empty Coordinates Pruning}: Refine $\Tilde{\y}_{t,C}$ by pruning candidates that are not actual points based on $GT(\boldsymbol{\Tilde{y}}_{t, C})$, yielding $\y_{t, C}$.
    \item \textit{Latents Features Probability Estimation}: Use the \gls{squlpe}-F model to estimate $\boldsymbol{\mu}_{t}, \boldsymbol{\sigma}_{t}$ for the latents $\hat{\y}_{t}$.
    \item \textit{Latents Features Decoding}: Entropy decode $\hat{\y}_{t, F}$ from the \gls{esqh}-F bitstream using $\boldsymbol{\mu}_{t}, \boldsymbol{\sigma}_{t}$.
    \item \textit{Higher Resolution And/Or Quality PC Reconstruction}: Reconstruct the final \gls{pc} $\hat{\x}_{t}$ using the \gls{jpeg-pcc} decoder applied to $\hat{\y}_{t}$ followed by the SR if specified in the coding parameters.
\end{enumerate}

This process can be iterated as needed to generate multiple enhancement layers, with each new layer serving as the base for the subsequent one.

\subsection{Assumptions on the coding parameters}

To drive the design of the \gls{esqh} algorithm some assumptions on the coding parameters were introduced based on the most commonly used coding configurations for \gls{jpeg-pcc}. These assumptions serve to reduce the complexity of \gls{esqh} while maintaining its effectiveness. The key constraints are as follows:

\begin{enumerate}
\item \gls{esqh} and \gls{squlpe} were designed to handle scaling factors that are powers of 2. Scaling factors that are not powers of 2 can be used as coding parameters for \gls{jpeg-pcc}, but are rarely, if ever, considered.
Moreover, this constraint effectively corresponds to pruning the deepest levels of the octree representing the point cloud.
\item The ratio of scaling factors between consecutive layers is restricted to either $1$ (as in \gls{sqh}) or $2$. Indeed, a ratio $2^n$ can be achieved by applying \gls{esqh} $n$ times thus upscaling the latents by the correct factor. Although the theoretical ratio between scaling factors for consecutive rate points could be as large as the point cloud's resolution, ratios of 4 or greater are rarely used in practice due to the substantial disparity in rate and distortion values between source and target PCs. Furthermore, as evidenced in Fig.~\ref{fig:corr_matrix_sf_seq}, even a ratio of 2 significantly impacts latent alignment, suggesting that larger ratios would further complicate the estimation of probability distributions for target latents.

\end{enumerate}

These constraints are reasonable tradeoffs (as will be shown in Section~\ref{sec:rqulpe results}) that allow to effectively cover the most common coding configurations used in \gls{jpeg-pcc} while maintaining a manageable level of complexity in the design of the \gls{squlpe} model.

\subsection{Coordinates Coding}
The process of applying \gls{esqh} when $sf_s \geq sf_t$ begins with encoding the information necessary to decode the coordinates $\y_{t,C}$.
This is accomplished using \gls{squlpe}-C, which is trained to predict the occupancy probabilities for a given set of target voxels. 
By using transposed sparse convolutions, every source occupied voxel is subdivided into 8 different target voxels (equivalent to creating a new level for the octree), whose coordinates are given by:
\begin{equation}
\hat{\y}_{t, C} = \{2 \cdot \C + [i, j, k] \, | \, \C \in \y_{s, C}, [i, j, k] \in \{0, 1\}^3\}.
\end{equation}

\begin{figure}[t]
    \centering
    \includegraphics[width=\columnwidth]{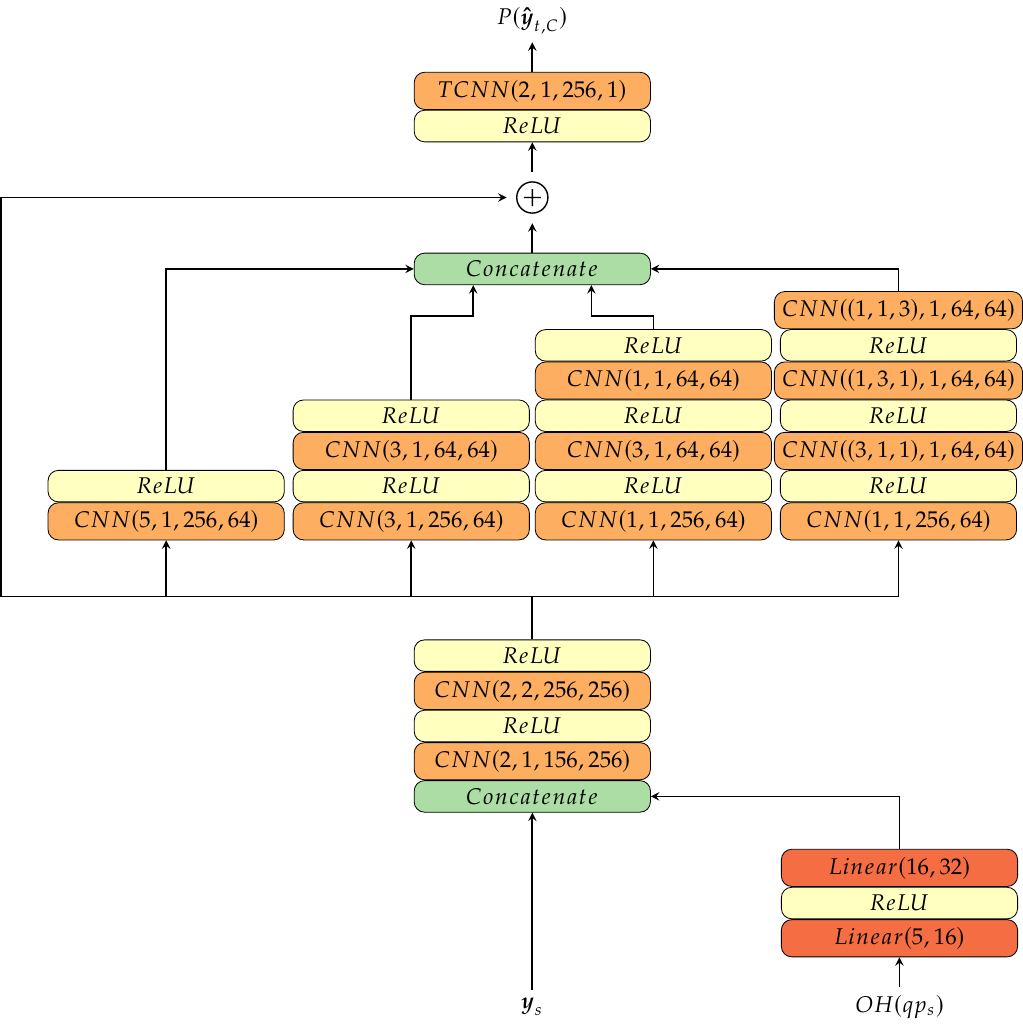}
    \caption{Scheme of the \gls{squlpe}-C network used to encode the latents coordinates losslessly.}
    \label{fig:p_predictor}
\end{figure}

Any of the resulting voxels could be empty or occupied in the higher resolution grid, and for this reason 
a ground truth label $GT(\C)$ is computed as: 
\begin{equation}
GT(\boldsymbol{c})
 =
\begin{cases}
    1 & \text{if } \C \in \y_{t, C} \\
    0 & \text{otherwise}
\end{cases}
\end{equation}

\begin{figure*}
    \centering
    \includegraphics[width=.8\textwidth]{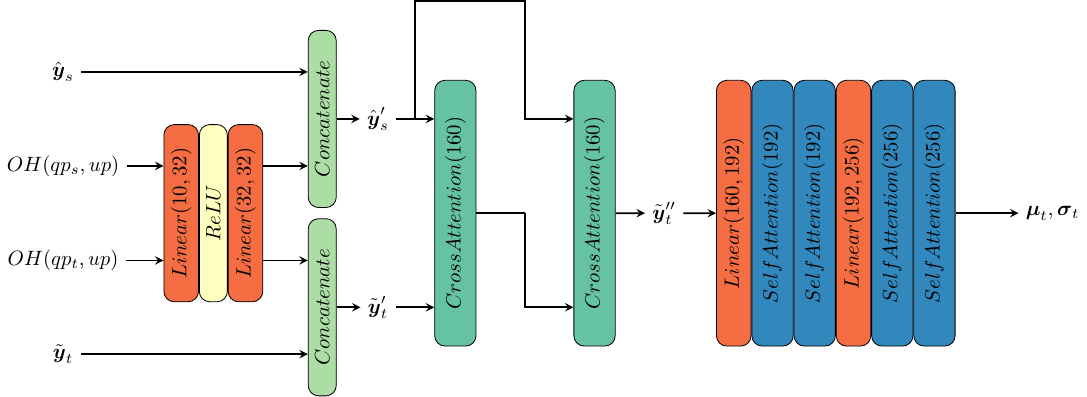}
    \caption{Scheme of \gls{squlpe}-F.}
    \label{fig:srqulpe}
\end{figure*}

At this point, the \gls{squlpe}-C model is trained using the cross-entropy loss between its predicted probabilities $P(\hat{\y}_{t, C})$ and the ground truths $GT(\hat{\y}_{t, c})$.
Then, at coding time, the estimated probability distribution $P(\hat{\y}_{t, C})$ serves as the entropy model to encode the ground truth $\tilde{\y}_{t, C} = GT(\hat{\y}_{t, c})$.

The \gls{squlpe}-C model, as illustrated in Fig.~\ref{fig:p_predictor}, is implemented as a sparse convolutional network that predicts $P(\hat{\y}_{t, C}|\hat{\y}_s, qp_s)$. It takes as input the source latents $\hat{\y}_s$ and the source quality parameter $qp_s$ which is transformed into an embedding by feeding its one hot encoding  
$OH(qp_s)$ into an MLP. The embedding is concatenated with $\hat{\y}_s$ and fed into the main network branch, consisting of two sparse convolutional layers and followed by an Inception-ResNet layer. The choice of a relatively simple architecture for \gls{squlpe}-C is justified by the negligible size of the bitstream required to encode $\y_{t, C}$ ($< 4\%$) compared to that needed for $\y_{t, F}$. More complex models, such as autoregressive ones, were deemed unnecessary for this task. This approach enables efficient encoding of the coordinate information necessary for upscaling while maintaining a balance between model complexity and performance.

\subsection{Features Coding}
Using low-quality latents as a base layer to encode higher-quality ones requires designing a model that can predict $P(\y_{t}| \hat{\y}_{s})$. In this case, this probability distribution is modeled as a Gaussian 
 $\mathcal{N}(\boldsymbol{\mu}, \boldsymbol{\sigma})$ as in \cite{minnen2018joint}. This requires a model, from now on referred to as \gls{squlpe}-F, that can handle latents with possibly different resolutions and/or different qualities.
The existing \gls{qulpe} model from \cite{mari2024point} proves inadequate for this task due to its sparse U-Net architecture, which constrains input and output resolutions to be identical. Two potential architectural approaches emerge:
\begin{enumerate}
    \item A dual-model approach, with one model that handles source and target latents obtained with different qualities and one model that handles latents with different resolutions.
This would result in a high number of total network parameters, which is already considerably high (22M in the \gls{qulpe} model \cite{mari2024point}).
    \item A unified architecture approach with a single model that handles all possible latents configurations. This allows reducing to the minimum the additional number of parameters required to achieve joint resolution and quality scalability due to effective usage of the learned parameters.
\end{enumerate}

Given these considerations, the development focused on a novel architectural approach that minimizes additional parameters while enabling joint resolution and quality scalability.
Motivated by recent results in many different fields including \gls{pc} coding, an architecture based on the attention layers proposed for \gls{pct} \cite{wu2022point} was adopted instead of sparse convolutions. 
The reason for this choice is that these architectures do not require matching input and output coordinates, which enables a single model to handle various combinations of resolutions and qualities simultaneously. As demonstrated in Section~\ref{sec:rqulpe results}, transformer-based models can achieve comparable or superior performance to sparse convolution models with a reduced overall parameter count, enhancing computational efficiency.

A \gls{pct} layer takes two sparse tensors as inputs. When these are identical then the layer is referred to as a self-attention layer, while if they are different (e.g., one represents the source latents while the other the target latents), then it is referred to as a cross-attention layer. Using cross-attention proves particularly effective since it allows using known information (the source latents) to improve the estimated probability distribution for the target parameters. A key feature of the cross-attention layer is its ability to consistently output a tensor with dimensions matching those of the second input in terms of point count and coordinates.

Under these considerations the steps taken by the \gls{squlpe}-F model (see the architecture in Fig.~\ref{fig:srqulpe}) to predict the probability of the target latents features $\y_{t, F}$ are:
\begin{enumerate}
    \item \textit{Coarse estimate for the target latents}: Generate an initial estimate $\Tilde{\y}_{t}$ of the target latents. In particular for each coordinate in $\y_{t, C}$ (which is already available through the previous lossless encoding step using \gls{squlpe}-C), the features of its nearest neighbors in $\hat{\y}_{s}$, denoted as $\boldsymbol{\kappa}_{y_s, F}$, are averaged to obtain $\Tilde{\y}_{t}$.
    \item \textit{Embedding of the coding parameters}: Embed the quality parameters $qp_s, qp_t$, and a boolean $up$, indicating if $\frac{sf_s}{sf_t} = 2$, into an embedding vector computed by feeding the one-hot encodings of the coding parameters into a shared multilayer perceptron (MLP), following the approach introduced in \cite{mari2024point}.
    \item \textit{Integration of the latents and the embedding}: Integrate $\hat{\y}_s$ and $\Tilde{\y}_{t}$ with the corresponding embeddings via concatenation to obtain $\hat{\y}_s^\prime$ and $\Tilde{\y}_{t}^\prime$.
    \item \textit{Refinement of the target latents estimate}: Refine the estimate $\Tilde{\y}_{t}^\prime$ through cross-attention with the source latents $\hat{\y}_s^\prime$ to obtain $\Tilde{\y}_{t}^{\prime\prime}$ 
    \item \textit{Prediction of the latents probability}: Process $\Tilde{\y}_{t}^{\prime\prime}$ through some self attention layers to predict the parameters $\boldsymbol{\mu}_t, \boldsymbol{\sigma}_t$ describing the probability distribution of the target latents
\end{enumerate}

\begin{figure*}
    \centering
    \includegraphics[width=.7\textwidth]{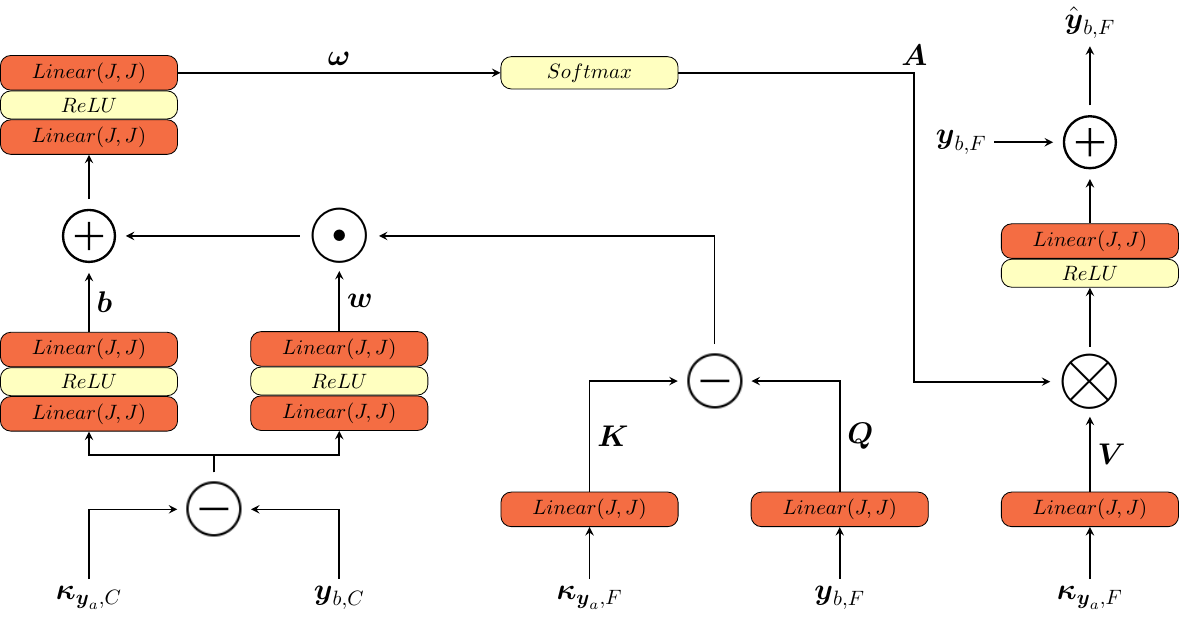}
    \caption{Scheme of the adopted attention layer.}
    \label{fig:pct}
\end{figure*}

This specific architectural choice enables the model to seamlessly handle scenarios where $\y_{s, C}=\y_{t,C}$ (implying $sf_{s}=sf_t$) as well as cases where $\y_{s, C}\neq\y_{t,C}$ (indicating $sf_{s} \neq sf_t$).

The next subsections will elaborate further on each specific step and their implementation details.

\subsubsection{\textbf{RQuLPE-F inputs}}
One of the inputs to the model is the initial estimation of the target parameters $\Tilde{\y}_{t}$ which was obtained by averaging the latents of the nearest neighbors. Alternative approaches for this initial estimation were explored, such as distance-weighted averaging schemes, these variations did not yield significant improvements in the final rate-distortion performance and were thus discarded.

As additional inputs the model takes the coding parameters $qp_s, qp_t, up$, however, while the quality parameters $qp_s$ and $qp_t$ are directly input to the model, the resolution information is simplified to the boolean $up$. This design choice allows for scalability across the most commonly used scaling factor ($sf$) values in \gls{jpeg-pcc}, which are typically powers of 2. The decision to handle only the case where $sf_s=2sf_t$ is based on the observation that larger ratios lead to poorly aligned latents, which the model struggles to exploit effectively. This suggests that employing multiple enhancement layers with a scaling factor ratio of 2 is a more efficient strategy than training a model to handle ratios exceeding 2.

The use of a boolean $up$ flag instead of explicit scaling factors is motivated by the understanding that the critical information is whether the latent resolutions differ, rather than their specific values. This is caused by the fact that point distributions (and consequently, latent representations) at various resolutions are heavily influenced by the original point distribution, which is unknown to the decoder. Therefore, explicit knowledge of $sf_s$ and $sf_t$ is unlikely to provide significant benefits and might potentially lead to overfitting.

\subsubsection{\textbf{Attention layers architecture}}

The attention architecture adopted in this work is depicted in Fig.~\ref{fig:pct}. The particular design of the layer allows for the exploitation of information from the first input $\y_a$ to enhance the representation of the second input $\y_b$. Through the computation of keys and values from $\y_a$ and queries from $\y_b$, the architecture enables the generation of a residual update for $\y_b$ based on a linear combination of values in $\y_a$.

In the architecture, $\K_{\y_a} = (\K_{\y_a,C}, \K_{\y_a,F})$ represents the nearest neighbors in $\y_a$ for each point in $\y_b$, with the neighboring relationship determined in the coordinate domain. While maintaining the core operations proposed in the original \gls{pct} work \cite{wu2022point}, this implementation extends beyond the original model by incorporating both self-attention and cross-attention mechanisms. Furthermore, it employs vector attention instead of grouped vector attention, with the number of selected groups set equal to the size of the vectors.

The choice of vector attention is particularly motivated by the nature of the latents produced by \gls{jpeg-pcc}, which are not computed using grouped vector attention. These latents lack specific structural organization along the channel dimension, making arbitrary channel grouping potentially suboptimal for overall performance. Therefore, processing each channel independently is considered more appropriate for this specific application, as it better aligns with the inherent characteristics of the \gls{jpeg-pcc} latent representations. This tailored approach to transformer architecture design enables more effective handling of \gls{jpeg-pcc} latents, potentially leading to improved RD performance.

A key design decision in all attention layers present in the \gls{squlpe}-F model is the use of 5 neighbors for the k-nearest neighbors (KNN) algorithm. This choice was taken after empirical testing revealed that increasing the neighbor count beyond 5 did not yield significant performance improvements but did increase computational complexity.

\subsection{Training and Evaluation}
The training and validation procedures for \gls{esqh} incorporate updates from the JPEG Pleno PCC Common Training and Testing Conditions \cite{jpeg-pleno-cttc}, reflecting modifications to both training and test sets to encompass a broader range of point cloud characteristics. These updated datasets, detailed in \cite{jpeg-pleno, jpeg-pleno-cttc}, were employed for training and evaluating both \gls{squlpe}-F and \gls{squlpe}-C components.

The latent representations were generated using \gls{jpeg-pcc} with quality parameters $qp \in {1, 2, 3, 4, 5}$ and scaling factors $sf \in {1, 2, 4}$ for each training and validation block. \gls{squlpe}-C and \gls{squlpe}-F are trained independently, as \gls{squlpe}-C's lossless coding objective for latent coordinates allows the use of ground truth during \gls{squlpe}-F training. Training and validation point clouds were segmented into $128\times 128\times 128$ blocks, an increase from the previous $64\times 64\times 64$ dimension, to better accommodate downscaling operations in conjunction with the intrinsic downscaling introduced by the analysis transform of \gls{jpeg-pcc}.

During training, at each gradient update step, a tuple ($qp_s, qp_t, sf_s, sf_t$)
is selected for each training PC block with uniform probability, and the
corresponding latents $\hat{\y}_s$, $\y_{t}$ are loaded accordingly from memory. 
This sampling strategy ensures comprehensive coverage of all possible configurations encountered during inference.
The validation phase implements an exhaustive evaluation across all parameter combinations, ensuring consistent validation loss measurements throughout the training epochs. 

Parameter combination selection differs between \gls{squlpe}-C and \gls{squlpe}-F, reflecting their distinct operational requirements:
\begin{itemize}
\item \gls{squlpe}-C: $qp_s \in {1, 2, 3, 4, 5}, sf_s = 2sf_t$, as the model specifically addresses latent resolution upscaling scenarios, disregarding $\y_{t, F}$.
\item \gls{squlpe}-F: $qp_s \leq qp_t + 1, 1\leq sf_s/sf_t \leq 2$, accommodating all permissible parameter combinations.
\end{itemize}

When training \gls{squlpe}-F, the quality parameter variations were constrained to $qp_s \leq qp_t + 1$ since values outside this range are rarely used in \gls{jpeg-pcc}.
While the condition $qp_s > qp_t$ can occur when $sf_s > sf_t$ (as downscaling often necessitates decreasing quality parameters for consecutive rate points), Fig.~\ref{fig:corr_matrix_sf_seq} indicates that the correlation between latents with different scaling factors and decreasing QPs is generally low. Preliminary tests revealed that training outside this range negatively impacts coding performance without providing substantial benefits.

\gls{squlpe}-C is trained with the loss function
\begin{equation}
    \mathcal{L}(\y_{t, C}, \hat{\y}_s )= \mathcal{H}(\y_{t, C}|\hat{\y}_s, qp_s)
\end{equation}

while \gls{squlpe}-F was trained to minimize
\begin{equation}
\mathcal{L}(\y_{t}, \hat{\y}_s) = \mathcal{H}(\y_{t}|\hat{\y}_s, \y_{t, C}, qp_s, qp_t, up)
\end{equation}

\begin{figure*}[!t]
    \centering
    \includegraphics[width=0.8\linewidth]{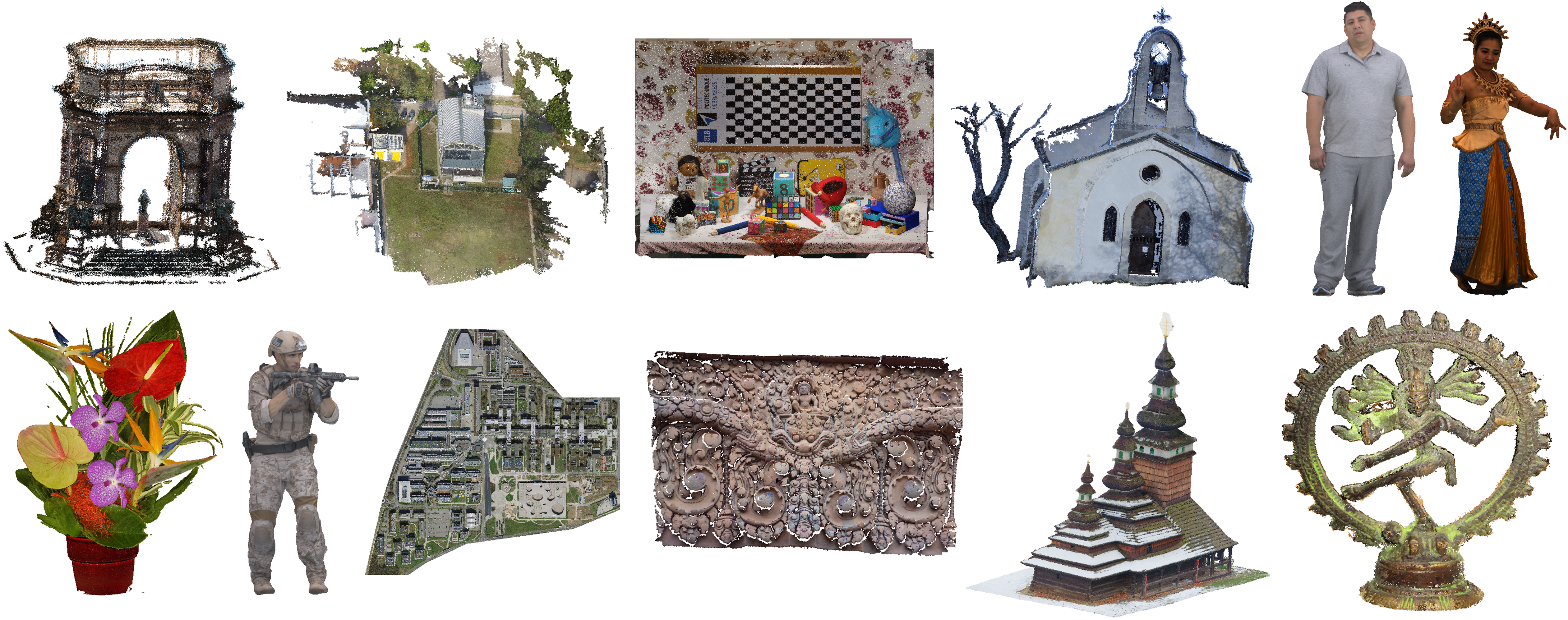}
    \caption{JPEG Pleno PCC test dataset. From left to right starting from the first row there are: Arco Valentino, CITIUSP, ULB Unicorn, House without roof, Boxer, Thaidancer, Bouquet, Soldier, EPFL, Facade 000009, Saint Michael, Shiva.}
    \label{fig:test_data}
\end{figure*}

The absence of a distortion component aligns with the objective of lossless encoding of $\y_t$ with minimal bit consumption. The optimization process utilizes the Adam optimizer with an initial learning rate of $10^{-3}$, implementing an exponential decay factor of 10 following 7 epochs without improvement. The training procedure incorporates early stopping, thus terminating when the validation error remains stagnant for 10 consecutive epochs.

%% file: 05_results.tex
\begin{table*}
    \caption{Coding parameters for JPEG PCC expressed as the tuple $qp, sf, SR$ where $SR \in \{T, F\}$ decides if the super-resolution module in \gls{jpeg-pcc} is used/not used.}
    \centering
    \begin{tabular}{|c|c|c|c|c|c|c|c|}
    \hline
    Type & PC & model & r1 & r2 & r3 & r4 & r5 \\
    \hline
    \multirow{8}{*}{Solid} & \multirow{2}{*}{StMichael} & \gls{jpeg-pcc} & 2, 4, T & 3, 2, T & 3, 1, F& 5, 1, F & \\
    &  & \gls{esqh} & 2, 4, T & 3, 2, T & 3, 1, F & 5, 1, F & \\
    \cline{2-8}
    & \multirow{2}{*}{Bouquet} & \gls{jpeg-pcc} & 5, 4, T & 3, 1, F  & 4, 1, F& 5, 1, F & \\
    &  & \gls{esqh} & 5, 4, T & \textcolor{red}{4, 2, T} & 3, 1, F  & 4, 1, F & 5, 1, F\\
    \cline{2-8}
    & \multirow{2}{*}{Soldier} & \gls{jpeg-pcc} & 3, 4, T & 3, 2, T & 3, 1, F& 4, 1, F & \\
    &  & \gls{esqh} &  3, 4, T & 3, 2, T & 3, 1, F& 4, 1, F & \\
    \cline{2-8}
    & \multirow{2}{*}{Thaidancer} & \gls{jpeg-pcc} & 2, 4, T & 2, 2, T & 2, 1, F& 4, 1, F & \\
    &  & \gls{esqh} &  2, 4, T & 2, 2, T & 2, 1, F& 4, 1, F & \\
    \hline
    \multirow{8}{*}{Dense} & \multirow{2}{*}{Boxer} & \gls{jpeg-pcc} & 1, 4, T & 2, 2, T & 5, 2, T & 4, 1, F & \\
    &  & \gls{esqh} & 1, 4, T & 2, 2, T & 5, 2, T& 4, 1, F & \\
    \cline{2-8}
    & \multirow{2}{*}{House wo roof} & \gls{jpeg-pcc} & 2, 4, T & 3, 2, T  & 4, 1, F& 5, 1, F & \\
    &  & \gls{esqh} & 2, 4, T & 3, 2, T  & 4, 1, F& 5, 1, F & \\
    \cline{2-8}
    & \multirow{2}{*}{CITIUSP} & \gls{jpeg-pcc} & 1, 4, F & 4, 4, F & 4, 2, F& 5, 2, T & \\
    &  & \gls{esqh} & 1, 4, F & 4, 4, F & 4, 2, F& 5, 2, T & \\
    \cline{2-8}
    & \multirow{2}{*}{Facade 00009} & \gls{jpeg-pcc} & 2, 4, T & 3, 2, T & 4, 1, F& 5, 1, F & \\
    &  & \gls{esqh} &   2, 4, T & 3, 2, T & 4, 1, F& 5, 1, F & \\
    \hline
    \multirow{8}{*}{Sparse} & \multirow{2}{*}{EPFL} & \gls{jpeg-pcc} & 1, 4, F & 4, 4, F & 4, 2, F& 5, 2, F & \\
    &  & \gls{esqh} & 1, 4, F & 4, 4, F & 4, 2, F& 5, 2, F & \\
    \cline{2-8}
    & \multirow{2}{*}{Arco Valentino} & \gls{jpeg-pcc} & 1, 4, F & 3, 4, F & 4, 2, F& 5, 2, F & \\
    &  & \gls{esqh} & 1, 4, F & 3, 4, F & 4, 2, F& 5, 2, F & \\
    \cline{2-8}
    & \multirow{2}{*}{Shiva} & \gls{jpeg-pcc} & 2, 4, F & 3, 2, F & 5, 2, T& 4, 1, F & \\
    &  & \gls{esqh} & 2, 4, F & 3, 2, F & 5, 2, T& 4, 1, F & \\
    \cline{2-8}
    & \multirow{2}{*}{ULB Unicorn} & \gls{jpeg-pcc} & 2, 4, F & 3, 4, F & 5, 4, F& 4, 2, F & \\
    &  & \gls{esqh} & 2, 4, F & 3, 4, F & 5, 4, F& 4, 2, F & \\
    \hline
    \end{tabular}
    \label{tab:coding params}
\end{table*}

\section{Performance Assessment for JPEG-PCC with SRQH}
\label{sec:rqulpe results}
This section presents the experimental setup and the performance assessment of the proposed \gls{esqh} implemented on top of \gls{jpeg-pcc}.
\subsection{Test Dataset}

The experimental evaluation utilized the JPEG Pleno PCC test dataset (Fig.~\ref{fig:test_data}), adhering to the Common Training and Test Conditions \cite{jpeg-pleno-cttc}. The test dataset comprises twelve point clouds, categorized into solid, dense, and sparse based on the MPEG-defined average local point density criteria, as detailed in Table~\ref{tab:coding params}.
\subsection{Coding Configurations}
The coding configurations utilized for coding the test dataset with \gls{jpeg-pcc} and \gls{squlpe} are documented in Table~\ref{tab:coding params}. These were derived for \gls{jpeg-pcc} by Guarda et.al. \cite{guarda2024jpeg} by optimizing the PCQM metric \cite{meynet2020pcqm} while meeting the target rates specified in the CTTC.
Regarding block size configuration, the coding configurations in Table~\ref{tab:coding params} were obtained with a block size of $128$. However, when using different SF parameters, since the blocks are partitioned after downscaling, the latents would correspond to different regions of the PC, thus preventing \gls{squlpe} from working correctly. For this reason, to provide a fair comparison between \gls{jpeg-pcc} and \gls{squlpe}, the block sizes used to code the PCs with \gls{jpeg-pcc}
were set to values equivalent to the ones enforced by \gls{esqh}. Specifically, the block size was defined as $128 \cdot sf_4/sf_1$, where $sf_1$ and $sf_4$ denote the scaling factors for the initial and final rate points respectively, ensuring a consistent block size of 128 at the final rate point.

An analysis of the configurations reveals that the vast majority of cases for \gls{jpeg-pcc} align with the constraints outlined in Section~\ref{sec:extending_sqh}. A single exception is observed for the \textit{Bouquet} point cloud, where the first and second rate points exhibit $sf_s=4, sf_t=1$ and $qp_s = qp_t + 2$, slightly deviating from the specified constraints.
To address this isolated case, the configurations for \gls{esqh} were adjusted to ensure full compliance with the established constraints, as reflected in Table~\ref{tab:coding params}. Importantly, this adjustment was implemented without compromising the integrity of the original coding configurations. The modification involved the inclusion of a single additional RD point to facilitate a smooth transition between two configurations with $sf_s=4, sf_t=1$, while all other RD points remained unaltered.

\begin{figure*}
    \centering
    \includegraphics[width=.8\textwidth]{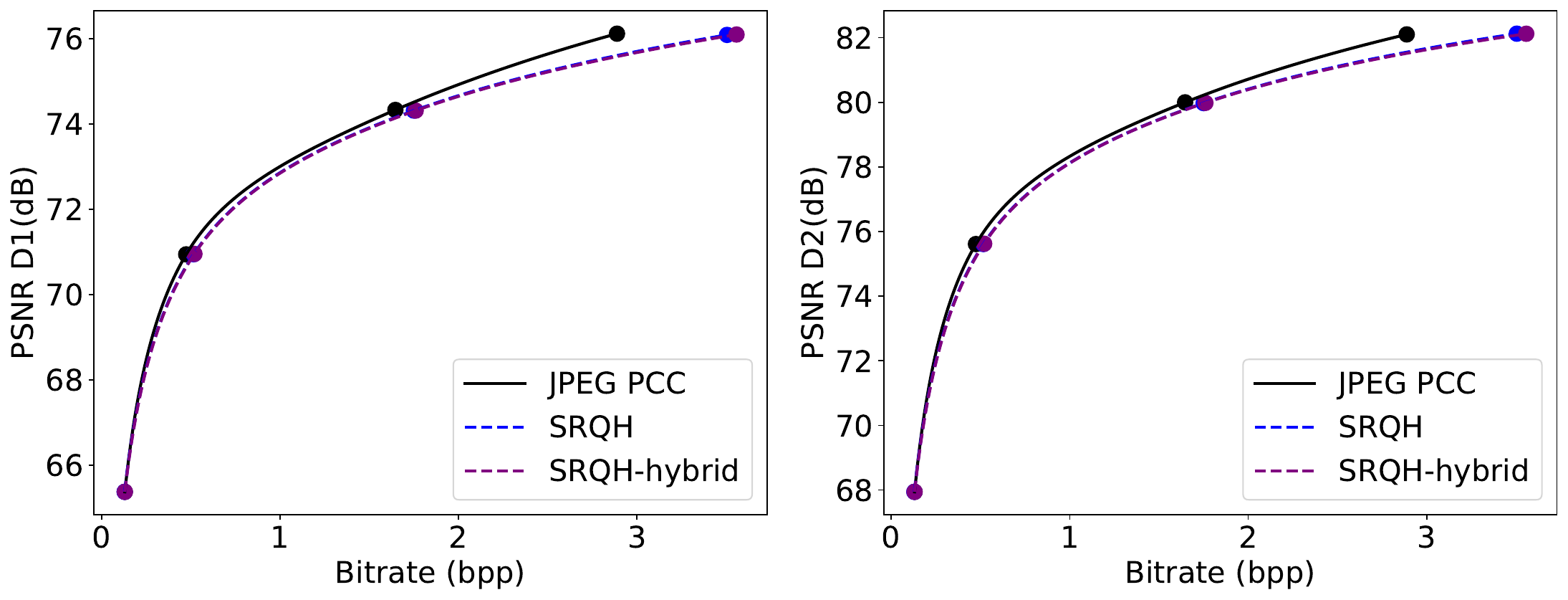}
    \caption{Average over the test set of the RD curves for \gls{jpeg-pcc} based solutions.}
    \label{fig:avg_jpeg}
\end{figure*}

This scenario underscores the robustness and flexibility of the implemented constraints. Their ability to accommodate the optimal configurations for the entire test set, with only minimal adjustments required in a single instance, demonstrates their practical applicability and effectiveness in real-world scenarios. The constraints thus prove to be sufficiently versatile to address the diverse requirements of point cloud coding across the test dataset.

\subsection{Metrics}
The assessment of decoded point clouds employed point-to-point PSNR (PSNR D1) and point-to-plane PSNR (PSNR D2) as quality metrics, while the bitrate was quantified using bits-per-(original)-point (bpp). The rate-distortion (RD) performance comparison against other anchors was evaluated using Bjontegaard delta rate and delta PSNR metrics.

\subsection{Baselines and Anchors}

The codec chosen as a baseline is \gls{jpeg-pcc} v4.0, i.e., the non-scalable codec serving as basis for the proposed \gls{esqh}

Additionally, another solution was evaluated, called \gls{esqh}-hybrid which is a hybrid scalable algorithm that selectively employs \gls{qulpe} for quality scalability ($sf_s = sf_t$) and \gls{squlpe} for resolution scalability ($sf_s \neq sf_t$).
This serves as an ablation study to investigate the potential advantages of utilizing specialized models for each scalability type versus a unified approach. For this comparative analysis, the \gls{qulpe} model underwent complete retraining using the updated training dataset.

Among the most widespread codecs, the ones that provide scalability for geometry coding are the conventional G-PCC and the learning-based SparsePCGC \cite{wang2022sparse} and Unicorn, \cite{wang2024versatile1,wang2024versatile2}
whose code and weights were unfortunately not publicly available when this work was developed.

For this reason, most of the codecs chosen as anchors do not provide any form of scalability. The chosen conventional anchors are: 
\begin{enumerate}
    \item G-PCC Octree v23.
    \item V-PCC Intra v24.
\end{enumerate}
while the chosen learning-based anchors are:
\begin{enumerate}
    \item GRASP-Net \cite{pang2022grasp}.
    \item PCGCv2 \cite{wang2021lossy}.
\end{enumerate}

Among these only G-PCC provides scalability (in particular resolution scalability) while all the other solutions are non-scalable.
To compare scalable and non-scalable solutions the RD curves for the scalable solutions were obtained by setting the quality metric for a specific rate point as the maximum possible quality obtainable with that bitstream.

\begin{table*}
    \centering
    \caption{Bjontegaard metrics of \gls{esqh} with reference to \gls{jpeg-pcc} based solutions.}
    \includegraphics[width=.9\textwidth]{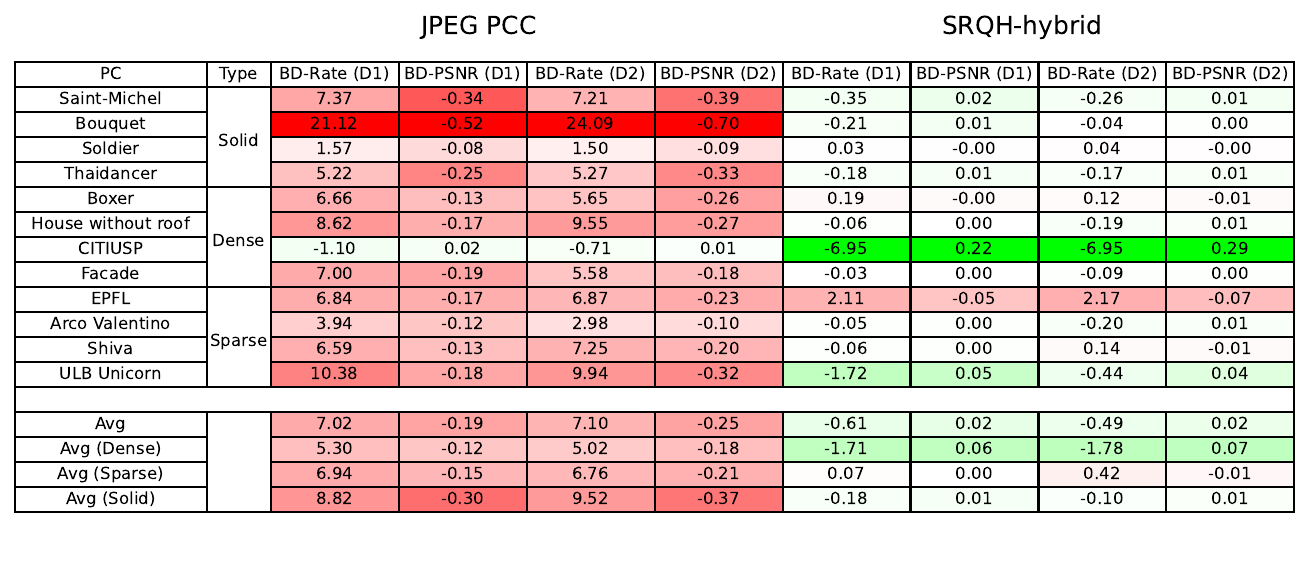}
    \label{fig:bd_rqulpe_jpeg}
\end{table*}
\begin{figure*}
    \centering
    \includegraphics[width=.8\textwidth]{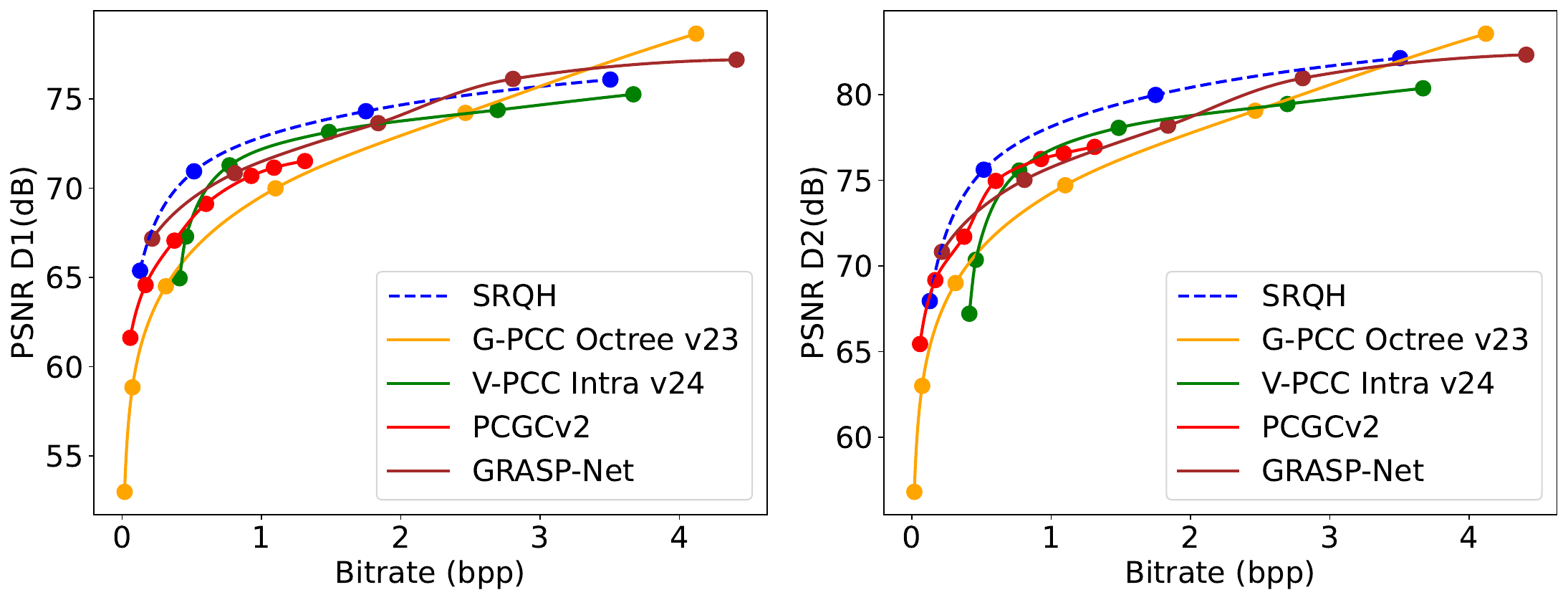}    
    \caption{Average RD curves when comparing against anchors that are not based on \gls{jpeg-pcc}.}
    \label{fig:rd_rqulpe_others}
\end{figure*}

\subsection{Performance Assessment}
This section presents the results obtained by the proposed solution against the baselines and the other anchors.

\subsubsection{\gls{esqh} versus JPEG-PCC-based coding solutions}

Firstly, in order to understand how much \gls{esqh} affects RD performance, it will be compared with non-scalable \gls{jpeg-pcc} and 
\gls{esqh}-hybrid. The RD curves obtained from the three solutions can be seen in Fig.~\ref{fig:avg_jpeg}
where their average over the whole test set is shown.

The results demonstrate that \gls{esqh} maintains performance comparable to \gls{jpeg-pcc}, with \gls{esqh} and \gls{esqh}-hybrid exhibiting nearly identical average performance. 
This achievement is particularly significant as it indicates that \gls{esqh} successfully implements full resolution and quality scalability while incurring minimal rate overhead compared to the non-scalable \gls{jpeg-pcc} solution. 

The Bjontegaard metrics in Table~\ref{fig:bd_rqulpe_jpeg} quantify the performance differences, revealing that the price to be paid for scalability is within 5-9\% across different PC categories which is acceptable for such an important feature. The most challenging case is observed for the \textit{Bouquet} point cloud, attributable to the fact that two consecutive configurations were not compliant with the constraints set before, since $sf_s/sf_t = 4$. This required adding an extra enhancement layer, thus increasing the price for scalability; additionally, the resulting configurations have $qp_s > qp_t$, thus a lower latent alignment.

The comparable performance between \gls{esqh} and \gls{esqh}-hybrid suggests that employing separate models for quality ($sf_s = sf_t$) and resolution ($sf_s \neq sf_t$) scalability offers no substantial benefits. In fact, \gls{esqh} slightly outperforms \gls{esqh}-hybrid on average, suggesting that exposure to diverse parameter configurations during training provides a beneficial regularizing effect. This finding favors the adoption of the proposed single unified \gls{squlpe} model, which achieves similar or slightly better rate-distortion performance with significantly lower computational complexity (4.7M parameters versus \gls{qulpe}'s 22M parameters). This additionally proves the effectiveness of \gls{pct} for this task which achieves similar performance with fewer network parameters than a much larger model based on sparse convolutions.

\begin{table*}
    \centering
    \caption{BD-Rate of \gls{esqh} with reference to G-PCC and V-PCC.}
    \includegraphics[width=.9\textwidth]{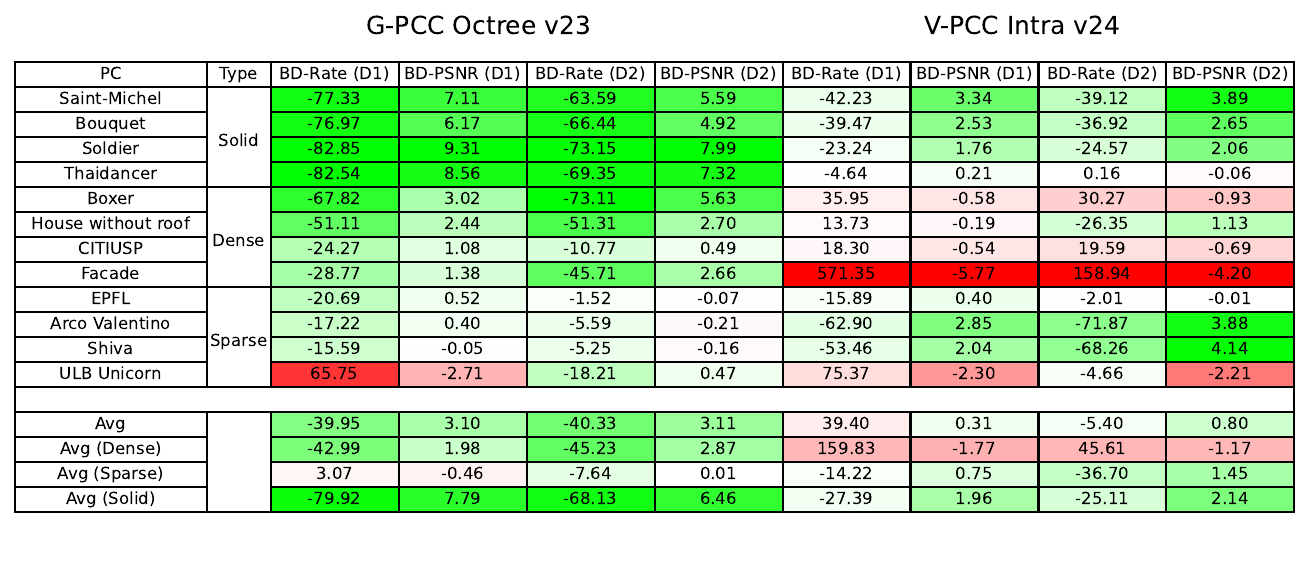}
    \label{fig:bd_rqulpe_standards}
\end{table*}

\begin{table*}
    \centering
    \caption{BD-Rate of \gls{esqh} with reference to GRASP-Net and PCGCv2.}
    \includegraphics[width=.9\textwidth]{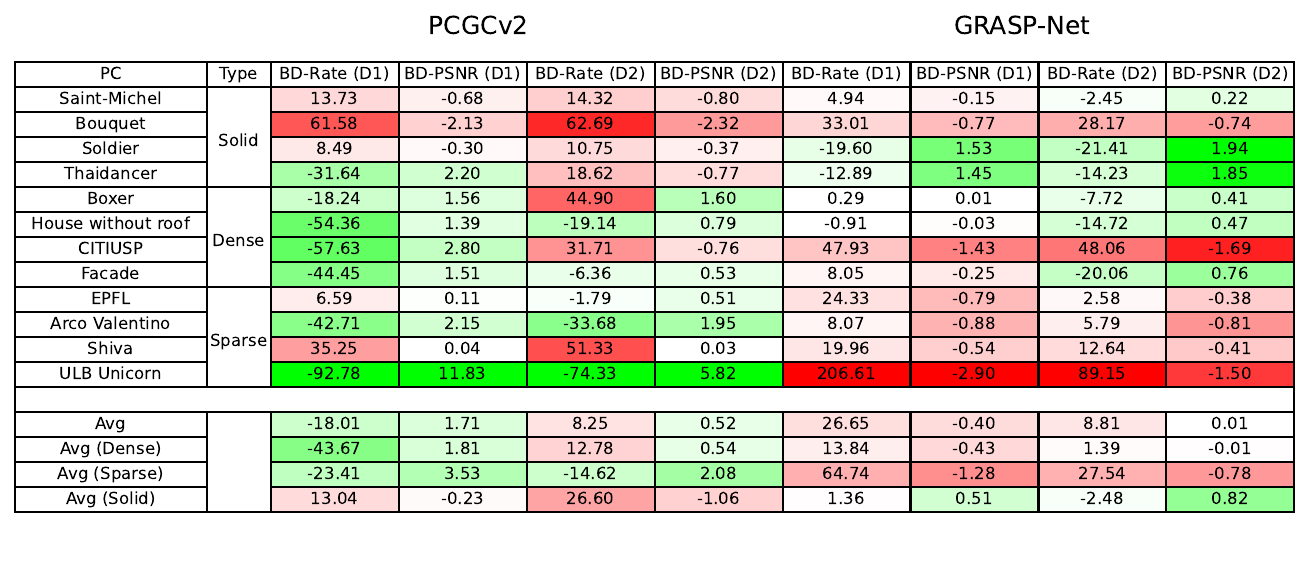}
    \label{fig:bd_rqulpe_others}
\end{table*}

As a final test \gls{esqh} was also compared with a naive scalable version of \gls{jpeg-pcc} where the bitstreams for each configuration in Table~\ref{tab:coding params} were concatenated together to achieve scalability. This solution yields 14.32\% higher bitrates on average (BD-rate D1) w.r.t. \gls{esqh} and 54\% additional rate when considering the full bitstream (i.e. the one relative to the highest rate point that allows decoding all chosen coding configurations). This shows that \gls{esqh} is an effective solution for providing scalability in \gls{jpeg-pcc}.




\subsubsection{\gls{esqh} versus anchors}

The performance comparison of \gls{esqh} against the anchor codecs, as illustrated in Fig.~\ref{fig:rd_rqulpe_others}, shows that despite being a scalable solution, \gls{esqh} performs better or on par with all the considered anchors. To better assess the relative performance between the solution and the anchors and to provide a more consistent evaluation, Bjontegaard-Delta (BD) metrics (BD-Rate, BD-PSNR) were employed, providing a more reliable basis for comparison despite their dependence on polynomial interpolation quality.

Table~\ref{fig:bd_rqulpe_standards} presents the BD metrics for \gls{esqh} in comparison with the standardized codecs (G-PCC and V-PCC), while Table~\ref{fig:bd_rqulpe_others} showcases the BD metrics for \gls{esqh} in comparison with the learning-based codecs (GRASP-Net and PCGCv2).
The analysis reveals that \gls{esqh} significantly outperforms G-PCC on solid and dense PCs, while showing slightly inferior performance on sparse PCs, primarily due to G-PCC's exceptional results on ULB Unicorn. In comparison with V-PCC, \gls{esqh} demonstrates superior performance on solid and sparse PCs, though V-PCC achieves better RD performance (D1 PSNR) on all four dense PCs in the test set.

Regarding learning-based codecs, \gls{esqh} outperforms PCGCv2 on dense and sparse PCs, with marginally lower performance on solid PCs. GRASP-Net shows comparable performance on solid PCs and superior results on dense and sparse PCs, indicating its effectiveness as an enhancement layer addressing G-PCC's limitations.

A critical distinction to note is that all compared methods, except G-PCC, lack scalability. In practical scenarios where scalability is required, \gls{esqh} offers a significant advantage since it enables decoding point clouds at multiple resolution/quality levels from a single bitstream.

\subsection{Complexity Analysis}

The computational impact of \gls{squlpe} model operations in scalable coding was assessed through comprehensive encoding and decoding time measurements. The evaluation was conducted on hardware comprising a single L40s GPU and 4 cores of an AMD EPYC 9224 CPU, with each PC processed 20 times, excluding extreme measurements to eliminate outliers.
To better assess the computational complexity, the PCs were coded with $qp \in \{1\dots5\}$ since coding PCs at different scales leads to very different coding times.
To obtain interpretable complexity metrics, testing focused on quality parameter variations ($qp \in {1, 2, 3, 4, 5}$) at original resolution ($sf=1$) with a block size of 128, excluding super-resolution effects to avoid accounting for postprocessing (which is equal for both \gls{jpeg-pcc} and \gls{esqh}) in the complexity evaluation.
This configuration was applied to both old and new test datasets, providing a robust sample size for complexity assessment. 

The evaluation framework measured four key temporal metrics:
\begin{itemize}
    \item $t_{enc, jpeg}(i)$: \gls{jpeg-pcc} encoding time, for non-scalable streams, required to encode the PC at quality $i$.
    \item $t_{enc, SRQH}$: \gls{esqh} encoding time for the full scalable bitstream.
    \item $t_{dec, jpeg}(i)$: \gls{jpeg-pcc} decoding time, for non-scalable streams, required to decode the PC at quality $i$.
    \item $t_{dec, SRQH}(i)$: time required by \gls{esqh} to decode the PC at quality $i$ from the scalable stream.
\end{itemize}

The relative computational overhead introduced by \gls{esqh} was quantified through
\begin{equation}
    t_{enc, extra} = \Bigg(\frac{t_{enc, SRQH}}{\sum_{i=1}^{5} t_{enc, jpeg}(i)} - 1\Bigg) \cdot 100 \\ 
\end{equation}
\begin{equation}
    t_{dec, extra}(i) = \Bigg(\frac{t_{dec, SRQH}(i)}{t_{dec, jpeg}(i)} - 1\Bigg) \cdot 100
\end{equation}

A key distinction exists between encoding and decoding processes: encoding requires full PC decoding at each rate point for distortion optimization, while decoding necessitates processing only the base layer and relevant enhancement layers without reconstructing the PC at each quality. 
For this reason, if quality $i$ needs to be decoded then the base layer and $i-1$ enhancement layers need to be decoded.
Consequently, \gls{esqh} execution time increases linearly with quality level $i$, suggesting a linear relationship between decoding time and enhancement layer count.

\begin{figure}
    \centering
    \includegraphics[width=.8\columnwidth]{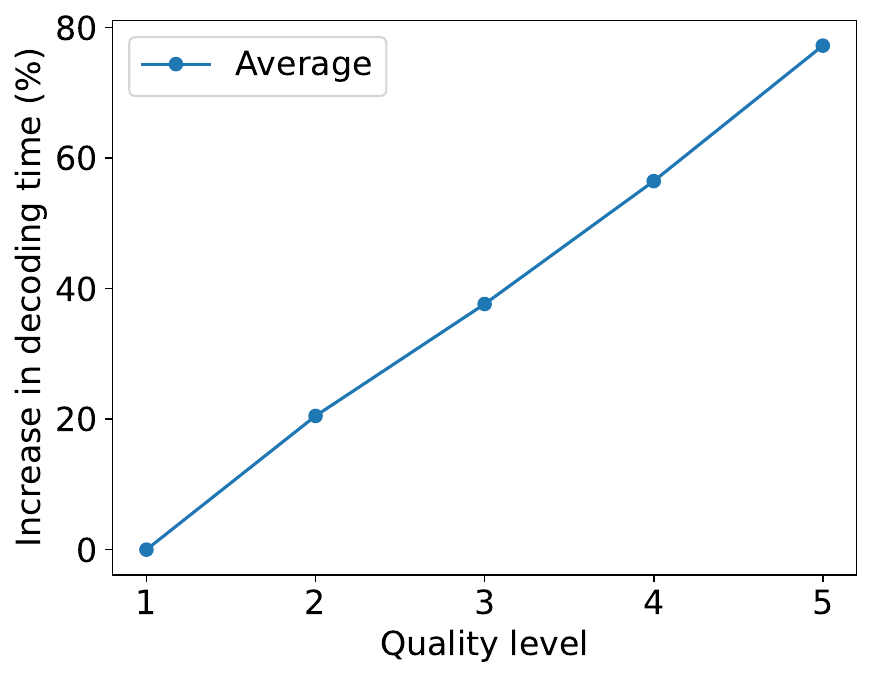}
    \caption{Extra decoding time, in percentage, required by \gls{esqh} w.r.t. \gls{jpeg-pcc}.}
    \label{fig:extra_dec}
\end{figure}

Experimental results showed that $t_{enc, extra} = 9.95\%$, additionally, the decoding time analysis, visualized in Fig.~\ref{fig:extra_dec}, confirms the expected linear growth with quality level, where each enhancement layer processing adds approximately 20\% to the base \gls{jpeg-pcc} decoding time. This value is higher than the average increase in encoding time since, 
at the encoder side, the time for distortion optimization is not negligible and it reduces the relative impact of \gls{esqh} w.r.t. \gls{jpeg-pcc} (a similar effect would be seen also if SR was introduced).

%% file: 06_conclusions.tex
\section{Conclusions}
\label{sec:conclusions}
This paper introduces \gls{esqh}, a novel joint resolution and quality scalability scheme implemented and validated for geometry coding in the \gls{jpeg-pcc} standard. \gls{esqh} enables the encoding of point clouds into scalable bitstreams, supporting compressed representations at various compression qualities and resolutions.
The proposed method demonstrates minimal to no rate-distortion performance loss compared to non-scalable \gls{jpeg-pcc}, indicating that the added scalability functionality incurs a negligible rate cost.

A key advantage of \gls{esqh} is its implementation in the latent space, which circumvents common drawbacks associated with spatial domain scalability, such as residual sparsity and the necessity to decode point clouds at each target quality. This approach results in limited additional complexity relative to the JPEG PCC baseline. Moreover, \gls{esqh} enhances the capabilities of \gls{sqh} by incorporating resolution scalability alongside quality scalability while reducing the required network parameters. The modular nature of \gls{esqh}, implemented through the \gls{squlpe} model, allows for flexible usage of the feature. Users can still utilize \gls{jpeg-pcc} in a non-scalable manner when scalability is not required, maintaining backward compatibility and versatility.

The latent space alignment principle underlying \gls{esqh} is readily achievable through sequential training, making this approach adaptable to other learning-based codecs.

Future research directions include integrating \gls{esqh} with JPEG-AI to enable attribute domain scalability in \gls{jpeg-pcc} and exploring additional applications of latent alignment beyond scalability. Preliminary findings suggest that low-quality latents may serve as superior side information compared to hyper-latents in certain scenarios, potentially leading to improvements in current entropy models. Additionally, extending support for arbitrary block sizes across various enhancement layers would enhance the algorithm's adaptability in practical applications.